\documentclass[10pt,journal,compsoc]{IEEEtran}

\usepackage{times}
\usepackage{epsfig}
\usepackage{graphicx}
\graphicspath{{fig/}}
\usepackage{amsmath}
\usepackage{amsthm}
\usepackage{amssymb}
\usepackage{dsfont}
\usepackage{algorithmic}
\usepackage{color}
\usepackage{soul}
\usepackage{caption}
\usepackage{tabularx}
\usepackage{booktabs}
\usepackage{bbm}

\usepackage{overpic}
\usepackage{tikz}
\usetikzlibrary{calc}
\usetikzlibrary{fit}

\usepackage[pagebackref=false,breaklinks=true,letterpaper=true,colorlinks,bookmarks=false]{hyperref}

\newif\ifdraft
\drafttrue

\definecolor{orange}{rgb}{1,0.5,0}
\definecolor{violet}{RGB}{70,0,170}

\ifdraft
 \newcommand{\PF}[1]{{\color{red}{\bf PF: #1}}}
 
 \newcommand{\DK}[1]{{\color{blue}{\bf DK: #1}}}
 
 \newcommand{\MK}[1]{{\color{green}{\bf MK: #1}}}
 
 \newcommand{\LC}[1]{{\color{orange}{\bf LC: #1}}}
 
\else
 \newcommand{\PF}[1]{}
 
 \newcommand{\DK}[1]{}
 
 \newcommand{\MK}[1]{}
 
 \newcommand{\LC}[1]{}
 
\fi

\newcommand{\comment}[1]{}

\newcommand{\Loss}{L}
\newcommand{\MSE}{\textnormal{\textsc{MSE}}}
\newcommand{\TOPO}{\textnormal{\textsc{TOPO}}}
\newcommand{\LMSE}{\Loss_\MSE}
\newcommand{\LTOPO}{\Loss_\TOPO}

\newcommand{\maxD}{D_\textnormal{max}}
\newcommand{\y}{y}
\newcommand{\f}{f}
\newcommand{\x}{x}
\newcommand{\dmap}{D}
\newcommand{\ydist}{\y_\dmap}
\newcommand{\p}{p}
\newcommand{\q}{q}
\newcommand{\rr}{r}
\newcommand{\I}{I}

\newcommand{\maximin}{\textnormal{maximin}}
\newcommand{\distance}{d}

\newcommand{\dmaximin}{\distance_\maximin}
\newcommand{\Road}{\mathcal{R}}

\newcommand{\BackgroundParcels}{\mathcal{B}}
\newcommand{\A}{A}
\newcommand{\B}{B}
\newcommand{\w}{w}
\newcommand{\vv}{v}
\newcommand{\connectivity}{\textnormal{conn}}
\newcommand{\disconnectivity}{\textnormal{disc}}
\newcommand{\Lconn}{\Loss_\connectivity}
\newcommand{\Ldisc}{\Loss_\disconnectivity}

\newcommand{\RTracer}{{\it RoadTracer}}
\newcommand{\Segm}{{\it Segmentation}}
\newcommand{\SegPath}{{\it Seg-Path}}
\newcommand{\DRoad}{{\it DeepRoad}}
\newcommand{\RCNN}{{\it RCNNU-Net}}
\newcommand{\MultiB}{{\it MultiBranch}}
\newcommand{\LinkN}{{\it LinkNet}}

\newcommand{\PolyM}{{\it PolyMapper}}
\newcommand{\Dru}{{\it DRU}}

\newcommand{\OursGlobal}{{\it TOPO-global}}
\newcommand{\OursWindowed}{{\it TOPO-windowed}}
\newcommand{\UNet}{{\it U-Net}}

\newcommand{\DG}{{\it DeepGlobe}}

\newcommand{\CNL}{{\it Canals}}

\newcommand{\APLS}{{\it APLS}}
\newcommand{\TLTS}{{\it TLTS}}
\newcommand{\Junc}{{\it JCT}}
\newcommand{\HM}{{\it HM}}
\newcommand{\CCQ}{{\it CCQ}}

\begin{document}
	
	\author{Doruk~Oner,
		Mateusz~Kozi\'{n}ski, 
		Leonardo~Citraro, \\
		Nathan C. Dadap,
		Alexandra G. Konings,
		Pascal~Fua  
	\IEEEcompsocitemizethanks{
		\IEEEcompsocthanksitem D.Oner, M.Kozi\'{n}ski, L.Citraro and P. Fua are with the Computer Vision Laboratory, \'{E}cole Polytechnique F\'{e}d\'{e}rale de Lausanne.
}
	\IEEEcompsocitemizethanks{
		\IEEEcompsocthanksitem N.C.Dadap and A.G.Konings are with the Remote Sensing Ecohydrology Group, Departemnt of Earth System Science, Stanford University.
}
	}

	\title{Promoting Connectivity of Network-Like Structures by Enforcing Region Separation} 
	
	\IEEEtitleabstractindextext{
		
		\begin{abstract}
We propose a novel, connectivity-oriented loss function for training deep convolutional networks to reconstruct network-like structures, like roads and irrigation canals, from aerial images.
The main idea behind our loss is to express the connectivity of roads, or canals, in terms of disconnections that they create between background regions of the image. 
In simple terms, a gap in the predicted road causes two background regions, that lie on the opposite sides of a ground truth road, to touch in prediction.
Our loss function is designed to prevent such unwanted connections between background regions, and therefore close the gaps in predicted roads.
It also prevents predicting false positive roads and canals by penalizing unwarranted disconnections of background regions.
In order to capture even short, dead-ending road segments, we evaluate the loss in small image crops.
We show, in experiments on two standard road benchmarks and a new data set of irrigation canals, that convnets trained with our loss function recover road connectivity so well, that it suffices to skeletonize their output to produce state of the art maps.
A distinct advantage of our approach is that the loss can be plugged in to any existing training setup without further modifications.
\end{abstract}

	}
	
	\maketitle

\section{Introduction}

Reconstruction of road networks from aerial images is a classic computer vision problem~\cite{Bajcsy76a,Vanderbrug76,Quam78,Fischler81b}, which remains actively studied to this day~\cite{Cheng17,Mattyus17,Li18h,Bastani18,Batra19,Chu19,Mosinska20,Yang19}. 
By contrast, the reconstruction of drainage canals was, so far, out of focus of the vision community.
However, it is of practical importance for hydrologic analyses~\cite{Murdiyarso10,Leifeld19}, which are becoming even more crucial at times of rapid climate changes.
Due to their network-like structure, canals are amenable to reconstruction by the same algorithms as roads, and we address these two problems jointly.
Most of the existing approaches~\cite{Cheng17,Mosinska18,Batra19,Yang19} rely on convolutional networks to extract from images binary masks denoting which pixels belong to roads and which do not. Unfortunately, they do not guarantee that the connectivity of the produced masks corresponds to that of the real road network. 
This is because these methods are trained to minimize losses, such as cross-entropy and mean squared error, that do not explicitly enforce topological consistency. 
When the annotations do not perfectly coincide with the imaged structures, which is always the case of satellite image annotations, 
networks trained with the per-pixel losses produce binary masks plagued by topological errors, such as road interruptions, missed junctions, and false positive connections.

In recent literature, this problem has been addressed by combining a convolutional encoder with a decoder that represents a network of roads as a graph, as opposed to a binary mask~\cite{Bastani18,Li18h,Chu19}. At inference time, the graph is grown iteratively: At each step, the neural network adds a new node to the graph by taking image features and the current state of the graph into account.
By contrast to the approach based on representing a road map as a binary mask, these graph-based methods make it easy to prevent excessively penalizing predicted roads that deviate slightly from their ground truth models, and to account for existing connectivity when growing the graph.
However, the non-differentiability of the node insertion operation makes training these networks more difficult and brittle than training convnets.

\begin{figure}[!htb]
\setlength{\tabcolsep}{1pt}
\centering
\begin{tabular}{c c c }
\includegraphics[width=0.16\textwidth,]{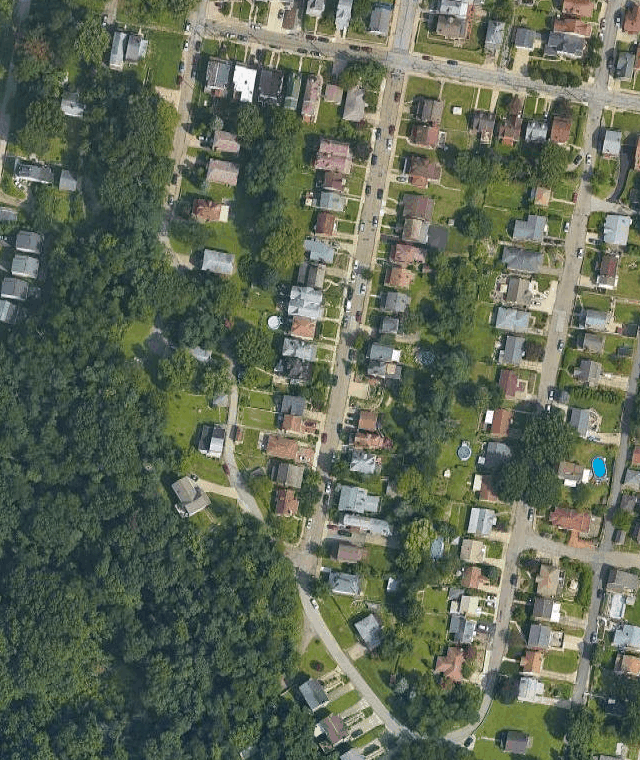} &
\includegraphics[width=0.16\textwidth]{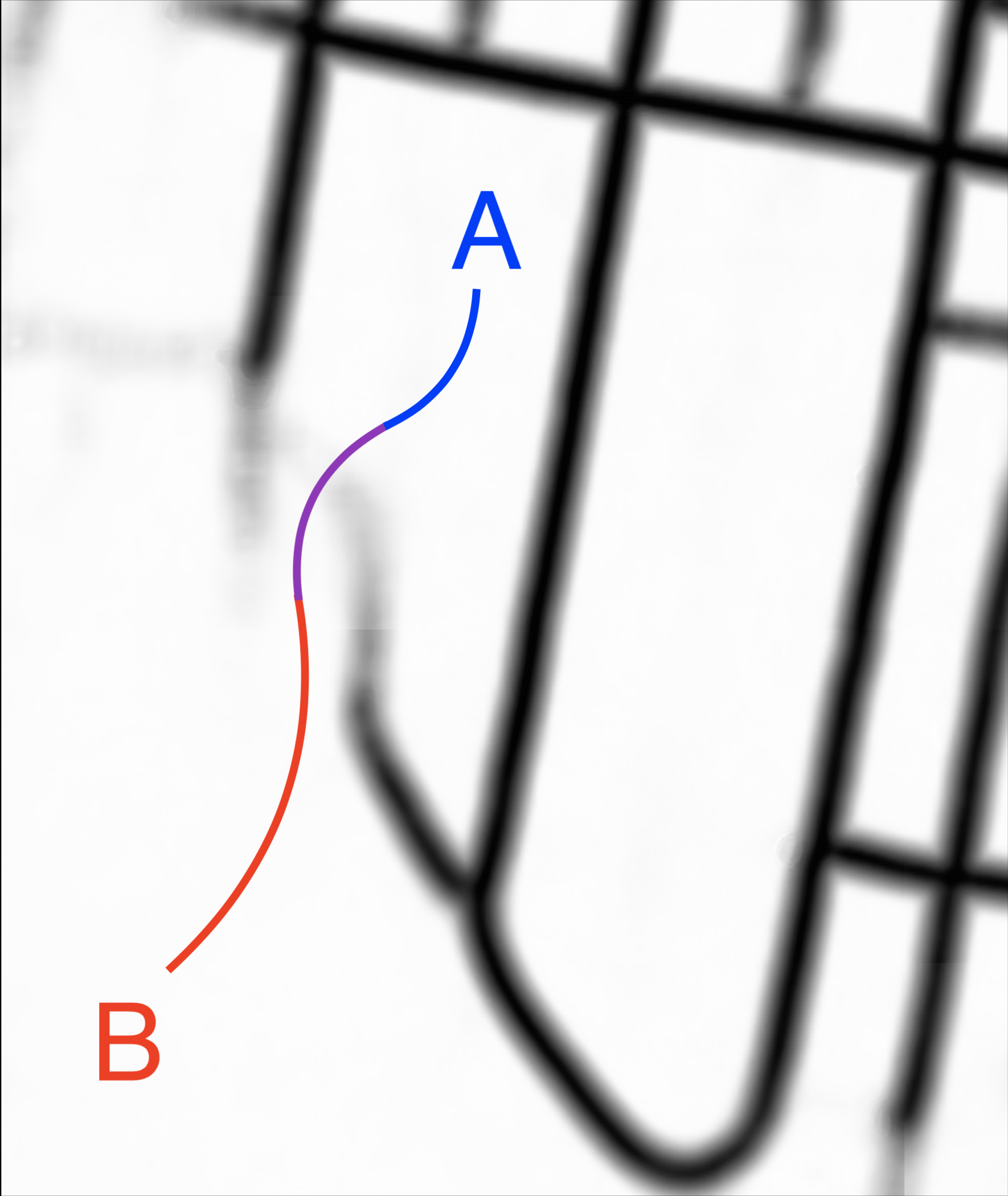} &
\includegraphics[width=0.16\textwidth]{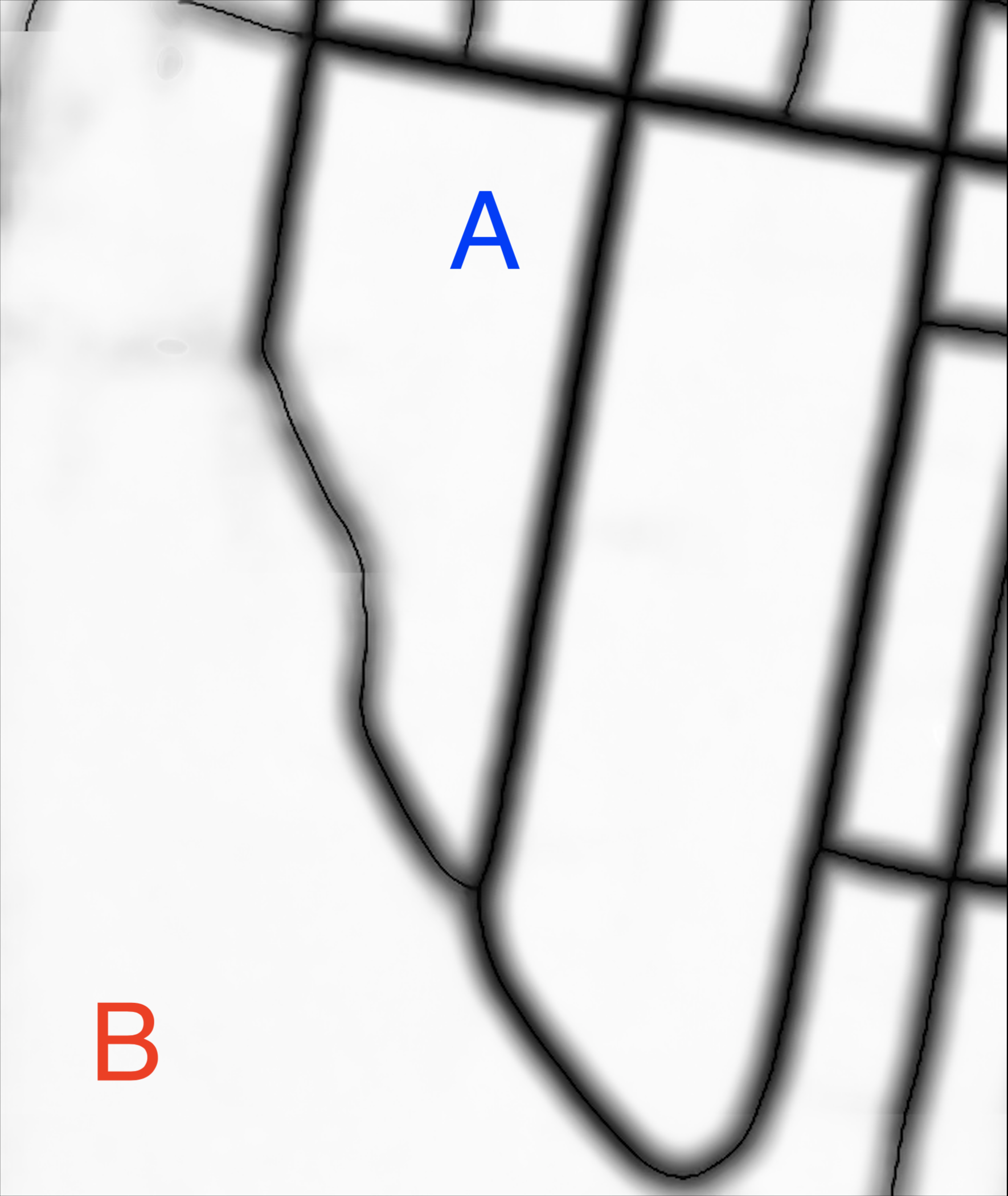} \\
\includegraphics[width=0.16\textwidth]{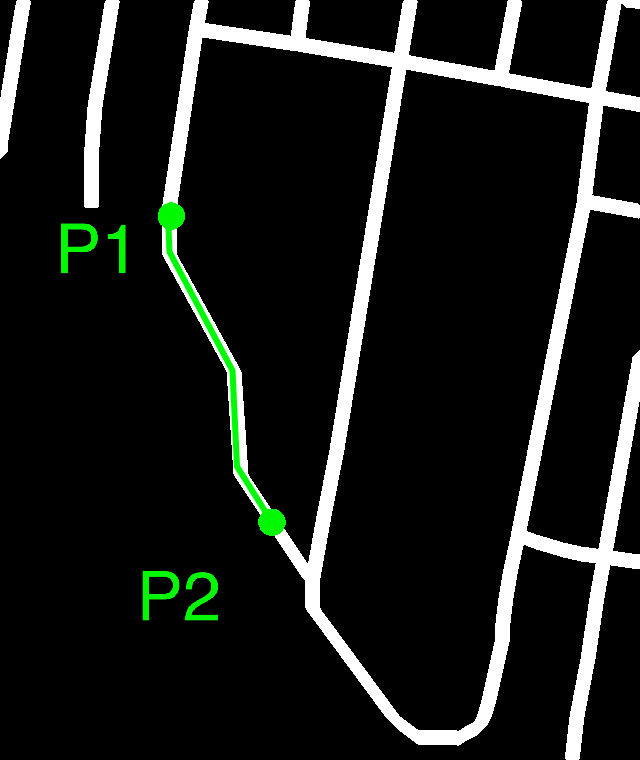} &
\includegraphics[width=0.16\textwidth]{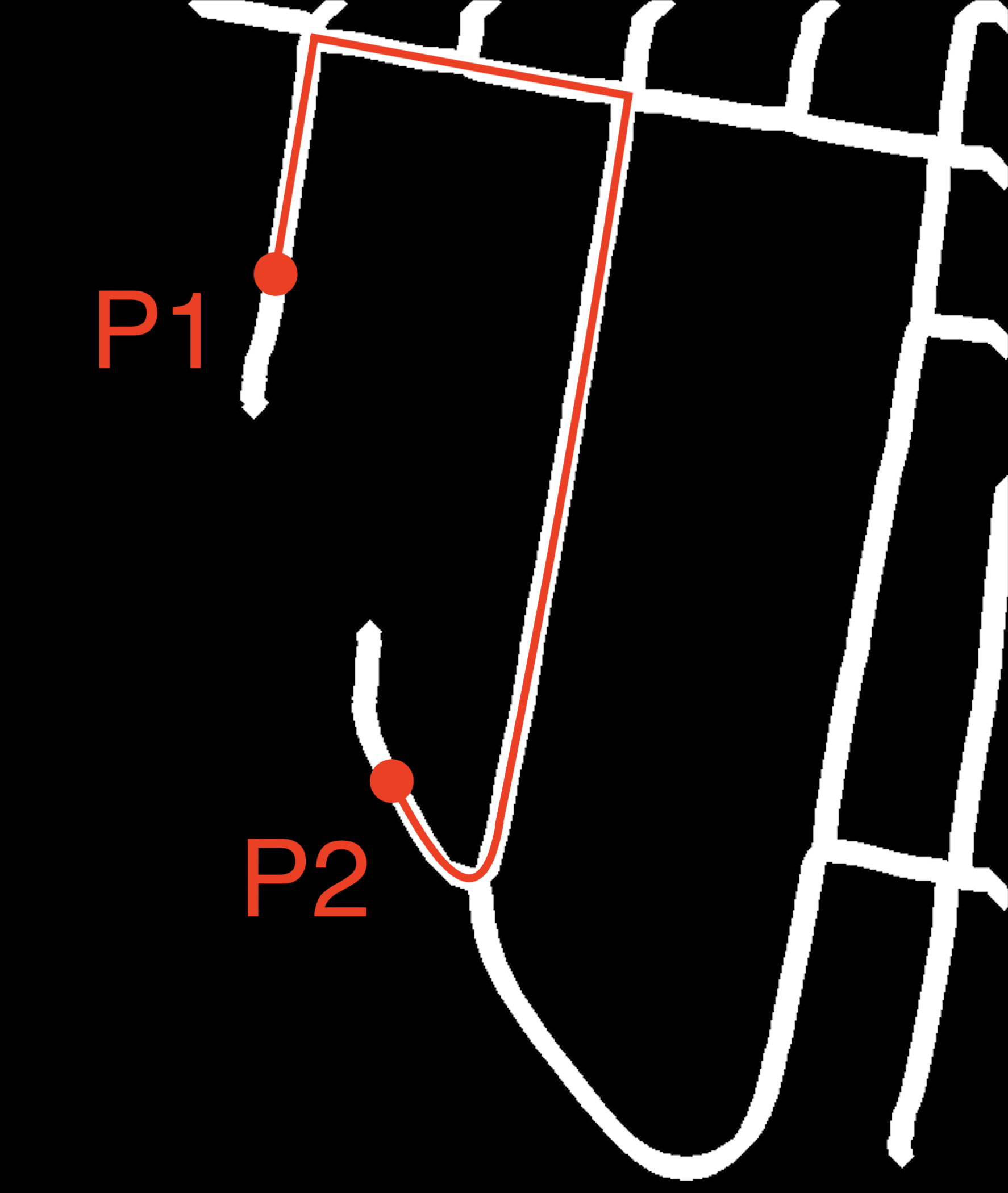} &
\includegraphics[width=0.16\textwidth]{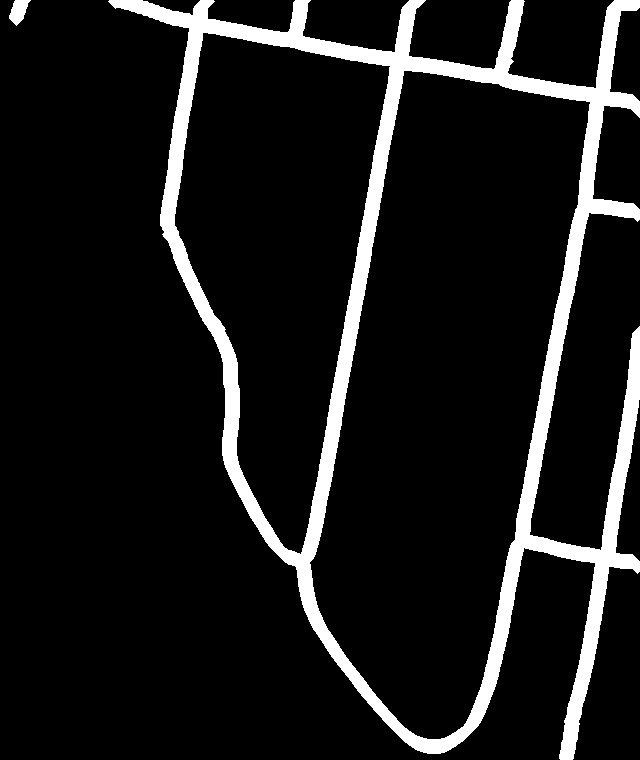} \\
(a)&(b)&(c)
\end{tabular}
\caption{{\bf We enforce road connectivity by penalizing connections between background regions.} (a) Input image and ground truth. (b) A distance map predicted by a U-Net trained {\it without} our connectivity loss, and its skeletonization, thickened for visibility.
Note that, even though there is a gap between road pixels $P_1$ and $P_2$, they remain connected both in the ground truth and in the prediction, because alternative paths exist in the loopy road network.
By contrast, background regions $A$ and $B$ connect in the prediction, but not in the ground truth.
(c) A distance map predicted by a U-Net trained using our disconnectivity loss its skeletonization. Our loss function penalizes connections between $A$ and $B$, preventing gaps in the predicted road.
}
\label{fig:teaser}
\end{figure}

In this paper, we show that connectivity of road and drainage canal networks can be enforced directly on a convolutional neural net, in a fully differentiable manner, and without the need to represent the graph explicitly. This allows end-to-end training and results in increased performance.
Our approach consists in relaxing the usual requirement of coincidence of annotated and predicted foreground pixels.
Instead, we require that predictions contain uninterrupted sequences of foreground pixels that can deviate by a few pixels from the ground-truth annotations.
This enforces connectivity while dealing with possibly imprecise annotations.

The difficulty is to express this requirement in the form of a differentiable loss function that can be used to train a deep network. 
The central idea of our approach is to forgo enforcing connectivity of the pixels annotated as centers of roads or canals, which may not coincide with true roads or canals.
Instead, we express the connectivity of the annotated structures in terms of the disconnections that they create between regions annotated as background.
More precisely, we require that two regions separated by a line in the ground truth, are also separated in the prediction.
As shown in Fig.~\ref{fig:teaser}, this effectively enforces continuity of the predicted road or canal networks.
By requiring that connected components of pixels annotated as background remain connected in the prediction, we prevent predicting false positive road or canal segments.
To capture dead-ending segments, we compute our loss in small image windows, which are likely to be subdivided even by short road and canal sections. 
To enforce the (dis-)connectivity of image regions, we re-purpose the differentiable machinery proposed in the MALIS segmentation algorithm~\cite{Turaga09,Funke18}.

Our contribution therefore is a novel approach to enforcing global connectivity of reconstructions of network-like structures from images. It can be used to boost the performance of {\it any} road delineation deep network that outputs a binary mask of road versus non-road pixels, without having to change the network itself. This is in stark contrast to the graph networks that do require changing both the network architecture and the training procedure. We demonstrate on both roads and drainage canals, that a simple U-Net~\cite{Ronneberger15} trained with our loss function, and combined with a standard skeletonization algorithm, attains state of the art performance in terms of the connectivity of the reconstructed networks.

\section{Related Work}

The existing approaches to reconstruction of networks of drainage canals rely on dedicated sensing modalities, like multi-spectral imaging and lidar~\cite{Vernimmen20,Ishii16}, and require extensive user interaction.
We show that the canals can be reconstructed from visual spectrum satellite images and with little required correction, just like roads. 

Many existing road segmentation algorithms rely on convnets~\cite{Cheng17,Batra19,Mosinska18,Yang19,Mosinska20} and all of them face the same difficulty: Training them by minimizing a cross-entropy loss, which is a {\it local}, pixel-wise measure, does not guarantee that their output preserve the global connectivity of road networks. Training the network to multi-task and to find not only the road centerline but also its spatial extent~\cite{Cheng17,Batra19} or its orientation~\cite{Batra19} mitigates the problem but does not explicitly enforce better connectivity. 
We instead propose to explicitly define the loss function to evaluate the connectivity. 

\subsection{Connectivity-oriented loss functions}
Ours is not the first attempt to make a convnet capture connectivity of linear structures in images by incorporating connectivity-oriented terms in the loss function.
One existing approach, is to use a perceptual loss function that depends on the statistical differences between features computed by forwarding either the ground truth or the prediction through a pre-trained neural network~\cite{Mosinska18}.
While this loss is indeed non-local, and has been shown to improve the connectivity of the predictions, it does not model connectivity explicitly.
Instead, it heavily relies on the assumption, that a pre-trained neural network implicitly captures some topological properties of the input.
By contrast, our loss function models connectivity explicitly.

Loss functions explicitly evaluating the topology of the predicted masks have been proposed for medical image segmentation~\cite{Clough19,Hu19b}.
However, strictly topological techniques are focused on counting loops and connected components in the data, irrespectively of their spatial position,
and cannot distinguish between different branching patterns.
That makes these loss functions a good choice when the segmented object has a relatively simple topology, like the aortic valve,
but not well suited for roads, which exhibit complex branching patterns and form numerous loops.
Our loss function is intended for linear structures forming complex topologies, like roads.

\subsection{Connectivity-oriented neural architectures}
Problems with connectivity can be addressed by designing predictors that output graphs instead of per-pixel masks, and explicitly decide about the presence of connections between map nodes.
This can be done as a post-processing step by generating a pool of potential additional connections and  training a classifier to decide which of the candidates should be inserted into the network~\cite{Mattyus17,Mosinska20}. One drawback of this approach is that it is not end-to-end trainable. 

A more elegant alternative is to use graph neural networks to predict the road graphs directly from the images~\cite{Bastani18,Li18h,Chu19}. 
This approach has certain disadvantages. Inference consists in a sequence of non-differentiable node insertion operations, 
which makes such networks slower than convnets. They are also more difficult to train, because node insertion is conditioned on the current state of the graph, and heuristics are needed to decide what is the optimal operation when the graph built so far is inconsistent with the ground truth.
In our experimental evaluation, we show that a simple convnet can outperform these approaches when trained with our loss function and post-processed with a vanilla skeletonization.
However, we still think that predicting graphs from images has merit, and the idea could be applied on top of a convnet trained with our loss.

\subsection{Affinity learning}
\label{sec:malis}

To enforce region connectivity, we use the maximin formulation of MALIS~\cite{Turaga09,Funke18}, a connectivity-oriented approach to segmenting cells in electron microscopy images of neural tissue.
It relies on the observation, that the predicted strength of connection between a pair of pixels can be expressed as the lowest value that needs to be crossed when traveling between the pixels in the prediction.
If this value equals $\theta$, thresholding the prediction with $\theta'<\theta$ produces a connected component containing both pixels.
Thresholding the prediction with $\theta''>\theta$ breaks the connection between the pixels.
Formally, $\theta$ is called a maximin cost of a pixel pair, and MALIS incorporates it into a differentiable loss term which is maximized for all pairs of pixels that belong to the same annotated cell, and minimized for all pairs of pixels from different cells.

We could have used the same approach to enforce the connectivity of road or canal pixels in the output of a segmentation network. 
This would have been ineffective for two reasons. 
First, both roads and canals often form loops and even if a connection between two road pixels is missed, they may still be connected via a different path. As illustrated by Fig.~\ref{fig:teaser}, there is a gap between pixels $P_1$ and $P_2$. Yet they are still connected to each other. Hence, this disconnection cannot be fixed simply by enforcing connectivity of any road pixel pairs.
Second, road and canal annotations usually take the form of one-pixel-thick centerline delineations that are rarely precise. 
Strictly enforcing the connectivity of centerline pixels would confuse the network and negatively impact its precision.

\begin{figure*}[!htb]
\input{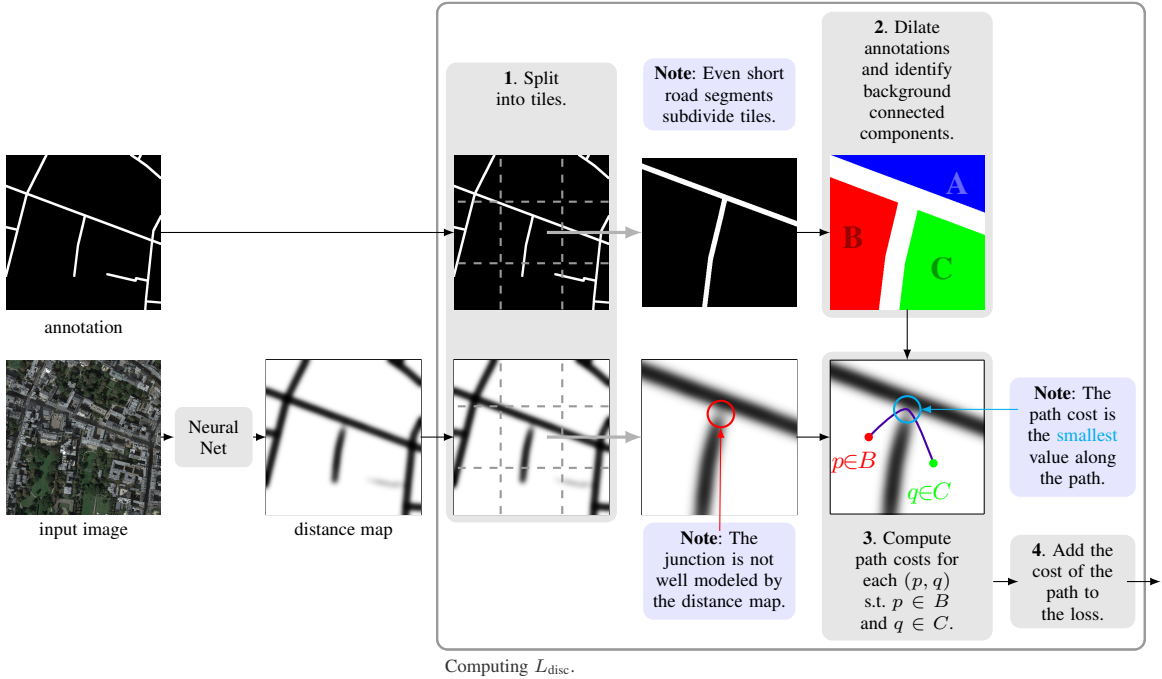}
\vspace{-3mm}
\caption{{\bf Computing $\Ldisc$.} We first tile the ground truth annotation and the distance map computed by our network (1). We use the ground-truth roads to segment each tile into separate regions (2). When there are unwarranted gaps in the distance map, there is a least one path connecting disjoint regions such that the the minimum distance map value along that path is not particularly small. We therefore take the cost of the path to be that minimum value (3) and we add to our loss function a term that is the maximum such value for all paths connecting points in the two distinct regions (4). This penalizes paths such as the one shown here and therefore promotes the road graph connectivity.}
\label{fig:method}
\end{figure*}

\section{Method}
\label{sec:method}

Given a training set of $N$ aerial images $\{x_i\}_{1 \leq i \leq N}$ and corresponding ground-truth binary masks $\{y_i\}_{1 \leq i \leq N}$ representing the roads or drainage canals in these images, we want to train a deep network $f_{\Theta}(\cdot)$, with weights $\Theta$, that takes an image $x$ as input and returns a distance map $\hat{y}$, consistent with the ground-truth. Our goal is to ensure that $\hat{y}$ represents the same connectivity as $y$. To this end, we minimize
\begin{align}
R(\Theta) &= \sum_i L\big(y_i,f_{\Theta}(x_i)\big) \; ,  \label{eq:risk} \\
\Loss(\y,\hat{\y}) & =\LMSE (\y,\hat{\y}) + \alpha \LTOPO (\y,\hat{\y}) \; ,  \label{eq:totalLoss}
\end{align}
with respect to the network weights $\Theta$. Here the loss function $\Loss$ is the sum of two terms $\LMSE$ and $\LTOPO$, and $\alpha$ is a parameter of the method that we set empirically using a validation set.
$\LMSE$ is a regression loss, used to train the network to predict the distance from each pixel to the center of the closest road or canal as in \cite{Sironi14}.
This lets us penalize the deviation of the predicted road center from its annotated position more gently than when using the more standard cross entropy.
Allowing for these deviations enables the connectivity-oriented $\LTOPO$ to force the network to predict uninterrupted roads and canals even if they do not coincide perfectly with the annotations.
We describe both terms below in more detail.

\subsection{Regression Loss: $\LMSE$}
We define $\LMSE$ as the Euclidean norm of the difference between the predicted distance map $\hat{\y}$ and the distance map $\ydist$ generated from the ground truth binary mask $\y$
\begin{equation}
\LMSE(\y,\hat{\y}) =  \sum_{\p \in \I} \big(\hat{\y}[\p] - \min(\ydist[\p],\maxD)\big)^2 ,
\end{equation}
where $X[\p]$ denotes the value of image $X$ at pixel $\p$, and $\I$ is the set of pixel indices in the input image. In practice, we found it advantageous to cap the ground truth distance at $\maxD=20$ pixels. Otherwise, the loss was strongly affected by large values of the distance map far away from foreground structures.

$\LMSE$ can be used by itself to train the network. As shown in Fig.~\ref{fig:teaser}, this gives good results in terms of per-pixel precision but the resulting binary masks feature many unwarranted interruptions. To prevent this, we now turn to the second term of Eq.~\eqref{eq:totalLoss}. 

\subsection{Connectivity Loss: $\LTOPO$}

$\LTOPO$ is the topology term that comprises the main contribution of this paper. 
Its purpose is to penalize in a differentiable manner unwanted interruptions and false connections in the output distance maps.
Instead of explicitly penalizing the interruptions and false connections of the foreground, we formulate our loss function in terms of connectivity of the background regions.
As shown in Fig.~\ref{fig:teaser}, an erroneous break in a predicted road causes two background regions, separated by a road in the ground truth mask $\y$, to touch in the distance map $\hat{y}$ produced by the network.
The first component of our loss, $\Ldisc$, penalizes such contacts.
Similarly, a false positive road divides a small crop of the predicted distance map into two background regions,
while the same crop of the ground truth distance map contains a single connected component of the background. 
Such errors are penalized by the second component of our loss, $\Lconn$.
The full  topological loss takes the form
\begin{equation}
\LTOPO (\y,\f(\x)) = \Ldisc\big(\y,\f(\x)\big) + \beta \Lconn\big(\y,\f(\x)\big) \;,
\label{eq:TOPO}
\end{equation} 
where $\beta$ is a parameter of the loss.
We introduce $\Ldisc$ and $\Lconn$ below.

\subsubsection{Maximin Dis-Connectivity.}
\label{sec:disc}
As illustrated in Fig.~\ref{fig:teaser}, in order to discourage interruptions of a predicted road,
we identify all pairs of background regions that the road separates in the ground truth,
and penalize connections between these regions in the predicted distance map.
To that end, we follow the \emph{maximin} approach of Turaga et al.~\cite{Turaga09}.
Intuitively, since the value of a road or canal pixel in a correct distance map should be small, and the background pixels should be large,
 two background pixels in an image can be considered connected if there exists a path of large-valued pixels between them.
The `strength' of this connection, can be evaluated as the value of the smallest pixel on the path with the largest smallest pixel connecting the end points.
Therefore, for each pair of pixels that are separated by a road or canal in the ground truth, $\Ldisc$ contains the `strength' of the connection between them.
As a result, minimizing $\Ldisc$ ensures the disconnectivity of regions on the opposite sides of roads and canals and, indirectly, improves the connectivity of roads and canals.

The detailed computation of $\Ldisc$ is depicted in Fig.~\ref{fig:method}.
We first dilate the centerline annotations by 5 pixels, which corresponds to the largest displacement between the image and the annotation that we have observed in our training data. 
We can therefore assume all the road pixels belong to this dilated region, which we denote as $\Road$ and which can also contain non-road pixels. 
Let $\BackgroundParcels$ be the set of background regions, that is, connected components in the remainder of the image.
Let us consider two pixels $\q\in\A$ and $\rr\in\B$ such that $\A,\B\in\BackgroundParcels$ and $\A\ne\B$.
Intuitively, $\q$ and $\rr$ lie on different sides of an annotated road. 
Since road pixels should receive low predictions,
a path $\pi$ connecting $\q$ and $\rr$ crosses a predicted road in the distance map $\hat{\y}$ if, for at least one point $\p$ along the path, $\hat{\y}[\p]$ is close to zero.
We therefore define the cost of path $\pi$ in the predicted distance map $\hat{\y}$ as $c(\pi,\hat{y})=\min_{\p\in\pi} \hat{\y}[\p]$, 
and measure the `connectivity' between background pixels $\q$ and $\rr$,
in terms of the maximin cost $\dmaximin(\hat{\y},\q,\rr)= \max_{\pi\in \Pi(\q,\rr)} c(\pi,\hat{\y})$, where $\Pi(\q,\rr)$ is the set of all paths connecting $\q$ and $\rr$. 
We enforce road connectivity by minimizing the maximin cost for all pairs of pixels that are separated by a road in the ground truth.
To that end, we define our connectivity-enforcing loss as
\begin{equation}
\Ldisc (\y,\hat{\y}) = \sum_{\A,\B\in\BackgroundParcels,\;\A\ne\B}\sum_{\q\in\A,\rr\in\B} \dmaximin(\hat{\y},\q,\rr)^2 \;.
\label{eq:maximin}
\end{equation}
%
When computed naively, the loss~\eqref{eq:maximin} requires summing costs over pairs of pixels, which would be computationally expensive.
However, Turaga et al.~\cite{Turaga09,Funke18} have shown that, because $\dmaximin(\hat{\y},\q,\rr)$ is equal to the value of the smallest pixel that has to be visited when traveling between $\q$ and $\rr$ in the prediction $\hat{\y}$, $\Ldisc$ can be computed efficiently as a sum over pixels, as opposed to pixel pairs, as
\begin{equation}
\Ldisc (\y,\hat{\y}) = \sum_{\p\in\Road} \w_\p \hat{\y}[\p]^2   \;,
 \label{eq:disc}
\end{equation}
where $\w_\p$ counts the pairs of pixels whose maximin cost is equal to $\hat{y}[\p]$.
Formally, we denote the maximin path between a pair of pixels $\q,\rr$ by $\pi(\q,\rr)$ and define
\begin{equation}
\w_\p=\sum_{\A,\B\in\BackgroundParcels,\;\A\ne\B}\sum_{\q\in\A,\rr\in\B} \mathbbm{1}\big[\p=\arg\min_{\rho\in\pi(\q,\rr)} \hat{\y}[\rho] \big]\;,
\end{equation}
where $\mathbbm{1}[\cdot]$ is the indicator function. 
The algorithm for computing the $\w_\p$'s is based on the Kruskal's maximum spanning tree algorithm, and we refer the reader to~\cite{Funke18} for details.
Note that, following~\cite{Funke18}, we constrain the computation of the loss $\Ldisc$ to the dilated road regions $\Road$.
This speeds up convergence in the early stages of the training, when path minima may be found far away from true roads.

\subsubsection{Penalizing False Connections}
\label{sec:conn}
We could take $\LTOPO$ to simply be $\Ldisc$ but we have observed that this results in many false positive road segments and that this behavior is difficult to counteract only by balancing the regression and connectivity losses with the coefficient $\alpha$ in Eq.~\eqref{eq:totalLoss}. 
To remedy this, we introduce another loss term that enforces connectivity of background regions, preventing false positive roads, as 
\begin{equation}\label{eq:conn}
\Lconn (\y,\hat{\y})= \sum_{\A\in\BackgroundParcels} \sum_{\p\in\A} \vv_\p (\hat{\y}[\p]-\ydist[\p])^2 \;,
\end{equation} 
where $\vv_\p$ is the number of pairs of pixels $\q,\rr\in\A$, for which $\p$ is the smallest pixel on the maximin path between $\q$ and $\rr$,
and is computed similarly to $\w_\p$,
and $\ydist[\p]$ is the value of the ground truth distance map at pixel $\p$.

\subsubsection{Introducing Sliding Windows.}
We can compute $\LTOPO$  as described above on the whole image.
However, when we do that, almost all pixels are assigned weights $\w=0$ in Eq.~\eqref{eq:disc} 
and a single road pixel gets a weight equal to the product of the size of the connected components that the road should separate.
This is because, in the presence of a an evident road interruption, all maximin paths go through this interruption.
This might seem desirable in theory, but in practice it makes learning unstable. Since only a small minority of pixels generate extremely large gradients,
no error signal is distributed among the remaining ones.

To overcome this problem we compute $\LTOPO$ independently for $64 \times 64$ image patches that cover the image, and sum the results.
This ensures that  at least one road pixel per window is taken into account and that its weight is not larger than $N^2/4$, where $N$ is the number of pixels in the window. 
As shown in Fig.~\ref{fig:method}, this also lets us handle dead-ending roads that do not separate the global map into disjoint areas.

\section{Experiments}

We now describe the dataset we have tested our approach on, the baselines to which we compare our results, and the metrics we used to assess the quality of the reconstructions. We then demonstrate that our new loss improves the results of networks that rely solely on conventional losses and substantially outperform recently proposed road reconstruction methods.

\subsection{Datasets}
We performed experiments on three publicly available datasets.
\begin{itemize}

\setlength\itemsep{3mm}

\item \RTracer{}. A recently published dataset of high-resolution satellite images covering urban areas of forty cities in six different countries~\cite{Bastani18}. Fifteen cities are used as a validation set. The ground truth is generated using OpenStreetMap.

\item \DG{}. Aerial images of rural areas in Thailand, Indonesia and India~\cite{DeepGlobe18}. The dataset comprises around 8500 images, 6200 of which are used for training, 1200 for validation and 1100 for testing. For a fair comparison to~\cite{Batra19}, we use the same split, consisting in 4695 training and 1530 test images.

\item \CNL{}. Aerial images of water drainage canals in rural areas of Malaysia~\cite{PlanetTeam17}. The dataset comprises a single large orto-photograph, 9768x10718 pixels large. 95\% of the image is used as training data and the rest is for testing.

\end{itemize}
Together, these datasets exhibit a very large variation of terrain type, which makes them an exhaustive benchmark for aerial road and drainage canal network reconstruction. 

\subsection{Baselines}
We compare the results of our algorithm to the following state-of-the-art methods.
\begin{itemize}

\setlength\itemsep{1.4mm}

 \item \Segm{}. A baseline algorithm from~\cite{Bastani18}, combining segmentation, thresholding, skeletonization, and conversion of the skeleton to a graph.
Road network reconstructions for the RoadTracer dataset were made available online by the authors~\cite{Mapster18}.
 
 \item \RTracer{}. Iterative graph construction where node locations are selected by a CNN~\cite{Bastani18}.
The road network reconstructions were released publicly by the authors~\cite{Mapster18}.
 
 \item \SegPath{}. A unified approach to segmenting linear structures and classifying potential connections~\cite{Mosinska20}.
The road network reconstructions were provided to us by the authors.
 
 \item \RCNN{}. Recursive image segmentation with post-processing for graph extraction~\cite{Yang19}.
The authors provided the probability maps.
 
 \item \DRoad{}. Image segmentation followed by post-processing focused on fixing missing connections~\cite{Mattyus17}.
The graphs for the RoadTracer dataset were published by the authors of this data set~\cite{Mapster18}. 
 
 \item \PolyM{}. Reconstructing a map by sequential construction of closed polygons~\cite{Li18h}. 
The graphs were provided to us by the authors.
 
 \item \MultiB{}. A recursive architecture co-trained in road segmentation and orientation estimation~\cite{Batra19}.
To obtain the road network reconstructions, we trained the network using the code published by the authors.
 
 \item \LinkN{}. An encoder-decoder architecture~\cite{Chaurasia17} co-trained in segmentation and orientation estimation~\cite{Batra19}.
We trained the network using the code made available by the authors.
 
 \item \UNet{}. Our own implementation of U-Net~\cite{Ronneberger15} trained with mean squared error.
 
 \item \Dru{} A recurrent U-Net iteratively refining segmentation output~\cite{Wang19c}, trained by us with the mean squared error. 
 
\end{itemize}

\subsection{Network architecture and training details}
We compare these baselines against three variants of our approach introduced in Section~\ref{sec:method}.

\begin{itemize}

\setlength\itemsep{1.4mm}

 \item \UNet{} + \OursGlobal{}. A U-Net, trained with our connectivity loss computed in the full image. 
 
 \item \UNet{} + \OursWindowed{}. A U-Net, trained with our loss computed in windows of size $64\times64$ pixels.
 
\item \Dru{} + \OursWindowed{}. A recurrent U-Net~\cite{Wang19c}, trained with the windowed version of our loss. 
\end{itemize}

In the \UNet{} experiments, we used the standard U-Net~\cite{Ronneberger15} architecture, with five blocks, each with three sequences of convolution-ReLU-batch normalization.
Max-pooling in $2\times2$ windows followed each of the blocks.
The initial feature size was set to $32$ and grew to $1024$ in the smallest feature map in the network.
We augmented the input data with vertical and horizontal flips and random rotations.

In the experiments with \Dru{}, we used a recurrent U-Net with the same architectural features that we used in \UNet{} experiments. There is a dual-gated recurrent unit in the bridge part of the network~\cite{Wang19c}. During training, we used three recurrent iterations. After each recurrent iteration, the output of the network is used as an additional channel to the input of the next iteration. For the first iteration, this additional channel is set to 0. During inference, we used the output of the second iteration which produced the best results.

We trained the network with the ADAM algorithm~\cite{Kingma15}, with the learning rate set to $1e-4$.
We set the mixing coefficients $\alpha=1e-4$ in Eq.~\eqref{eq:totalLoss} and $\beta=0.1$ in Eq.~\eqref{eq:TOPO}, empirically.
\subsection{Performance measures}
Comparing connectivity of road reconstructions is difficult, because the reconstructions rarely overlap with the ground truth, and often deviate from it significantly.
There seems to be no consensus concerning the single best evaluation technique in the existing literature -- we have found five different connectivity-oriented metrics in concurrently published recent work.
To provide exhaustive evaluation, we used all of them in our experiments.
\begin{itemize}
\setlength\itemsep{2mm}
 \item \APLS{} Average Path Length Similarity, defined as a aggregation of relative length difference of shortest paths between pairs of corresponding points in the reconstructed and predicted maps~\cite{APLS}.
 
 \item \TLTS{} Statistics of lengths of shortest paths between corresponding pairs of end points randomly selected in the predicted and ground-truth networks~\cite{Wegner13}. We report the fraction of paths where the relative length difference is within 5\%. 
 
 \item \Junc{} A junction score, evaluating the number of roads intersecting at each junction~\cite{Bastani18}. Consists of road recall, averaged over the intersections of the ground-truth and road precision, averaged over the intersections of the prediction. We report the corresponding F1 score.
 
 \item \HM{} Compares the sets of graph locations accessible by traveling away from randomly chosen pairs of corresponding points in both graphs~\cite{Biagioni12}. We report the corresponding F1-score.
 
 \item \CCQ{} To complement the connectivity-oriented evaluation, we also computed the most popular metric that measures spatial co-occurrence of annotated and predicted road pixels, rather than connectivity. The Correctness, Completeness and Quality are equivalent to precision, recall and intersection-over-union, where the definition of a true positive has been relaxed from spatial coincidence of prediction and annotation to co-occurrence within a distance of 5 pixels~\cite{Wiedemann98}. We report the Quality as our single-number metric.
 
\end{itemize}

\begin{table}[!h]
\centering
\caption{
Results of experiments on the \RTracer{} dataset~\cite{Bastani18}.
Our loss function makes even the simple \UNet{} attain state of the art performance.
Computing the loss in windows results in improvement of four out of five performance criteria.
\label{tab:results-roadtracer}
}
\begin{tabular}{@{} p {0.335\columnwidth} >{\centering\arraybackslash}p{0.7cm}>{\centering\arraybackslash}p{0.5cm}>{\centering\arraybackslash}p{0.5cm} >{\centering\arraybackslash}p{0.4cm} >{\centering\arraybackslash}p{0.002cm} >{\centering\arraybackslash}p{1.45cm} @{} } 
\cmidrule{2-7}

 & \multicolumn{4}{c}{Connectivity-oriented} && pixel-based \\

\cmidrule{2-5}
\cmidrule{7-7}

 Method &    \APLS{} &       \TLTS &      \Junc &         \HM &&        \CCQ \\
\cmidrule{1-7}
\Segm{} \cite{Bastani18}&
        62.5 &        33.0 &       78.2 &        69.4 &&        54.4 \\ 
\RTracer{} \cite{Bastani18}&
        59.1 &        40.6 &        81.2 &        70.5 &&        47.8 \\
\SegPath{} \cite{Mosinska20}&
        68.1 &        46.5 &        75.4 &        67.6 &&        54.0 \\
\RCNN{} \cite{Yang19}&
        48.2 &        18.4 &       75.9 &        68.8 &&        62.8 \\
\DRoad{} \cite{Mattyus17}&
        24.6 &         6.4 &      51.4 &        46.8 &&        43.6 \\
\PolyM{} \cite{Li18h}&
        61.3 &        31.5 &       80.0 &       53.7 &&        35.7 \\
\UNet~\cite{Ronneberger15}&
		66.3 &        40.0&        77.5 &        68.2 &&        59.3 \\ 
\cmidrule{1-7}
\UNet+\OursGlobal{}   &        72.5 &        46.3 & \textbf{84.7} &       70.3 &&        63.8 \\
\UNet+\OursWindowed{} & \textbf{75.8} & \textbf{49.7} &        82.8 &\textbf{76.0} && \textbf{68.6}\\
\midrule
\cmidrule{1-7}
\end{tabular}
\end{table}

\begin{table}[!h]
\centering
\caption{
Results of experiments on the \DG{} dataset.
Our loss function improves the performance of both \UNet{} and \Dru{} in terms of all the metrics,
with \Dru{} attaining the state-of-the-art performance.
\label{tab:results-deepglobe}
}
\begin{tabular}{@{} p {0.335\columnwidth} >{\centering\arraybackslash}p{0.7cm}>{\centering\arraybackslash}p{0.5cm}>{\centering\arraybackslash}p{0.5cm} >{\centering\arraybackslash}p{0.4cm} >{\centering\arraybackslash}p{0.002cm} >{\centering\arraybackslash}p{1.45cm} @{} }
\cmidrule{2-7}

 & \multicolumn{4}{c}{Connectivity-oriented} && pixel-based \\

\cmidrule{2-5}
\cmidrule{7-7}

 Method &    \APLS{} &       \TLTS &      \Junc &         \HM &&        \CCQ \\
\cmidrule{1-7}
\LinkN{} \cite{Batra19}&
            67.7 &       60.6  &       66.2  &       73.4  &&         77.2 \\
\MultiB{} \cite{Batra19}&
        70.8 &      65.2  &       71.1  &       75.6  && 79.4\\
\UNet{}~\cite{Ronneberger15} &     62.3    &       59.9     &  66.4  &       72.7      &&     68.8   \\
\Dru{} \cite{Wang19c}&
75.2 &       65.4  &       67.2  &       76.6  && 80.1\\
\cmidrule{1-7}
\UNet+\OursWindowed{}  & 75.2 &\textbf{69.8}&\textbf{71.2}&\textbf{79.8}&&
77.0 \\
\Dru{} + \OursWindowed{}  & \textbf{77.1}& 68.4 &\textbf{71.2}&79.6&&
\textbf{80.7} \\
\cmidrule{1-7}
\end{tabular}

\end{table}

\begin{table}[!h]
\centering
\caption{
Results of experiments on the \CNL{} dataset.
Our loss function boosts the performance of both \UNet{} and \Dru{} in terms of all the five metrics.
\label{tab:results-canals}
}
\begin{tabular}{@{} p {0.335\columnwidth} >{\centering\arraybackslash}p{0.7cm}>{\centering\arraybackslash}p{0.5cm}>{\centering\arraybackslash}p{0.5cm} >{\centering\arraybackslash}p{0.4cm} >{\centering\arraybackslash}p{0.002cm} >{\centering\arraybackslash}p{1.45cm} @{} }
\cmidrule{2-7}

 & \multicolumn{4}{c}{Connectivity-oriented} && pixel-based \\

\cmidrule{2-5}
\cmidrule{7-7}

 Method &    \APLS{} &       \TLTS &      \Junc &         \HM & &       \CCQ \\
\cmidrule{1-7}
\UNet{}~\cite{Ronneberger15} &     70.9   &      30.3     &  76.4  &       61.4      &&     81.4   \\
\Dru{} \cite{Wang19c}& 71.4 & 32.7  &  76.9  &  62.1  && 80.3\\
\cmidrule{1-7}
\UNet+\OursWindowed{}  & 76.3 & 35.7 & \textbf{79.3} & 63.4 && \textbf{85.1} \\
\Dru{} + \OursWindowed{}  & \textbf{78.2}& \textbf{43.1} & 78.8 & \textbf{67.0} &&
84.5 \\
\cmidrule{1-7}
\end{tabular}

\end{table}

\subsection{Comparative evaluation}

\begin{figure*}[t]
	\centering

	\begin{tabular}{@{} >{\centering\arraybackslash}m{0.30\textwidth} >{\centering\arraybackslash}m{0.30\textwidth} >{\centering\arraybackslash}m{0.30\textwidth} @{}}
		\includegraphics[width=0.30\textwidth]{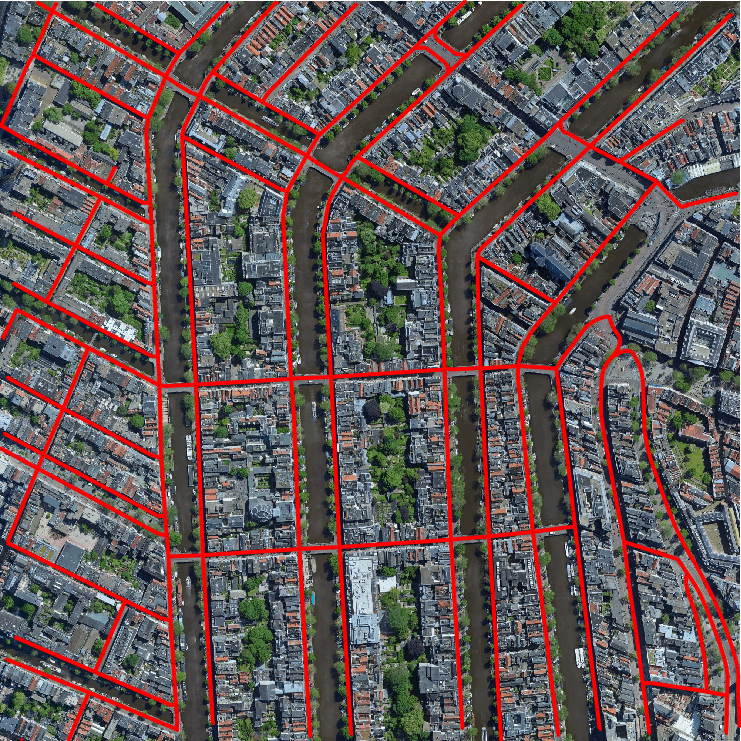}&
		\includegraphics[width=0.30\textwidth]{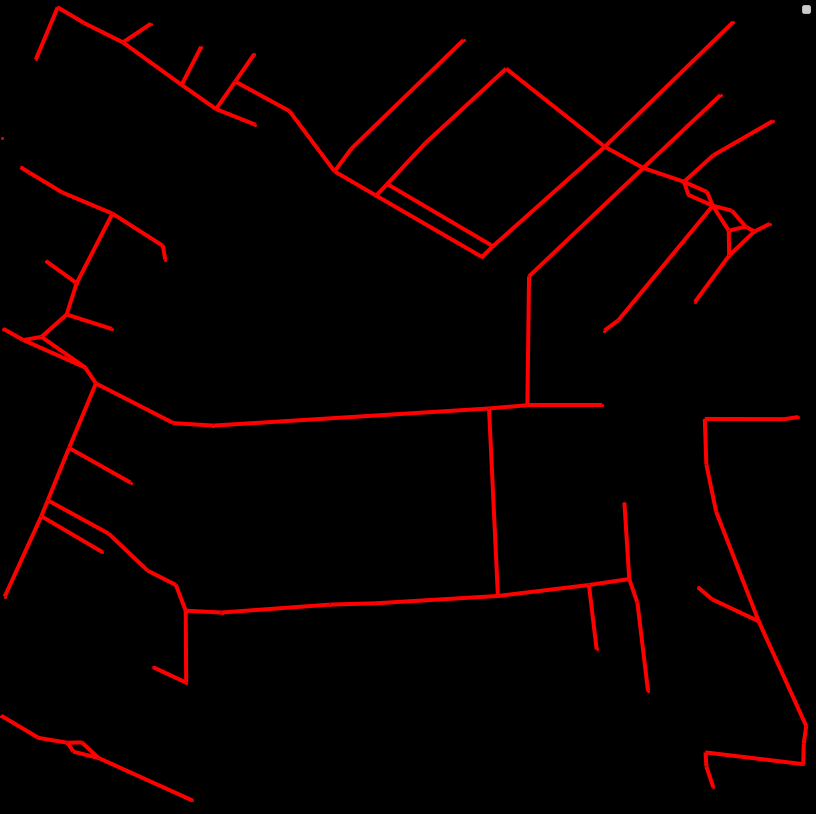} &
		\includegraphics[width=0.30\textwidth]{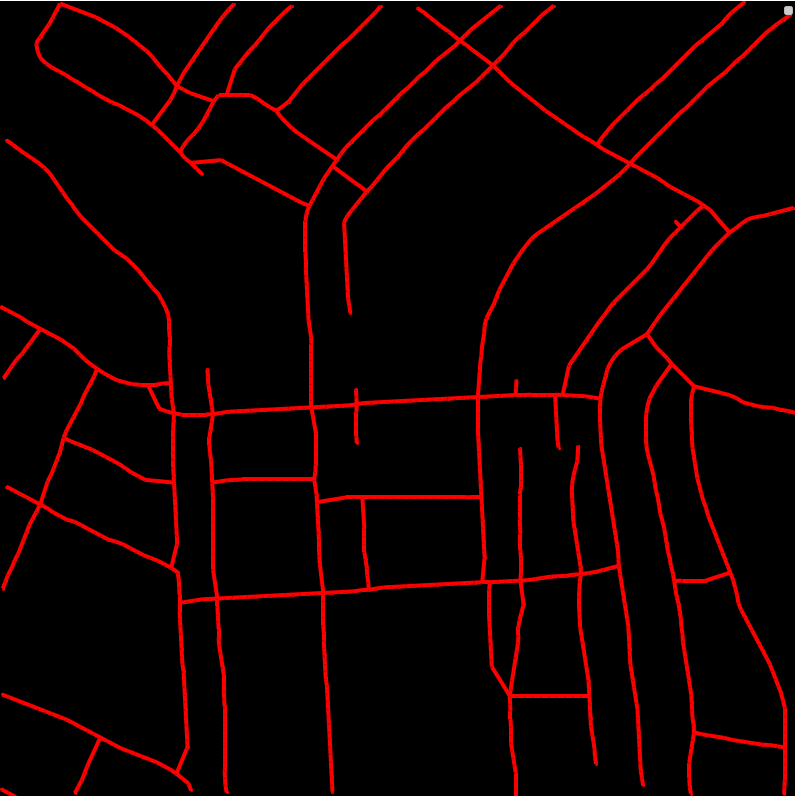} \\
		{\it input} &
		\Segm{} & 
		\RTracer{} \\
		
		\includegraphics[width=0.30\textwidth]{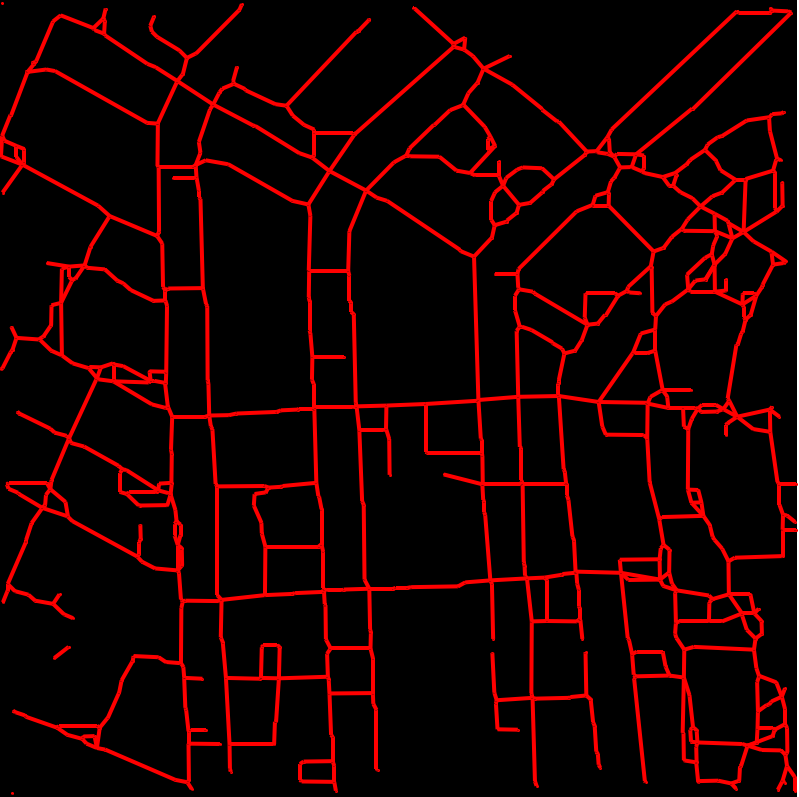} &
		\includegraphics[width=0.30\textwidth]{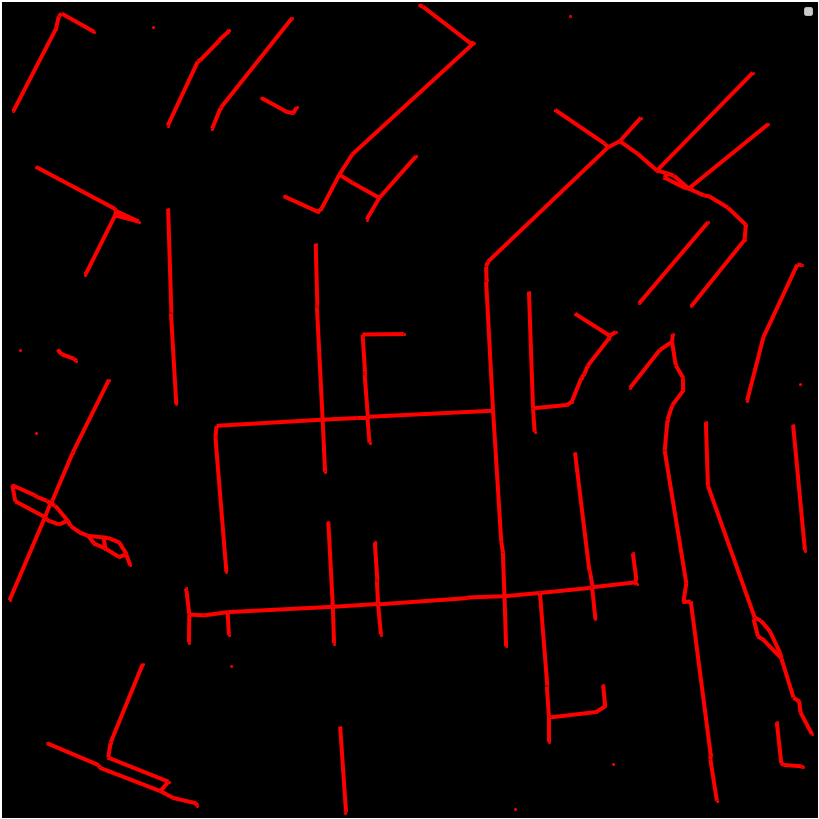} &
		\includegraphics[width=0.30\textwidth]{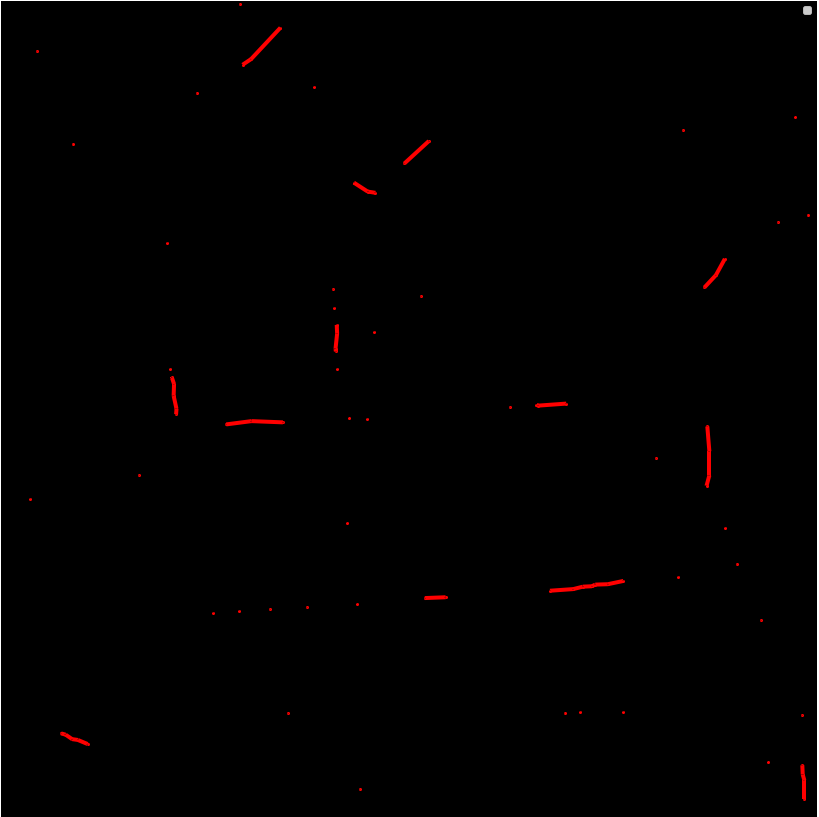} \\
		\SegPath{} &
		\RCNN{} &
		\DRoad{} \\
	
		\includegraphics[width=0.30\textwidth]{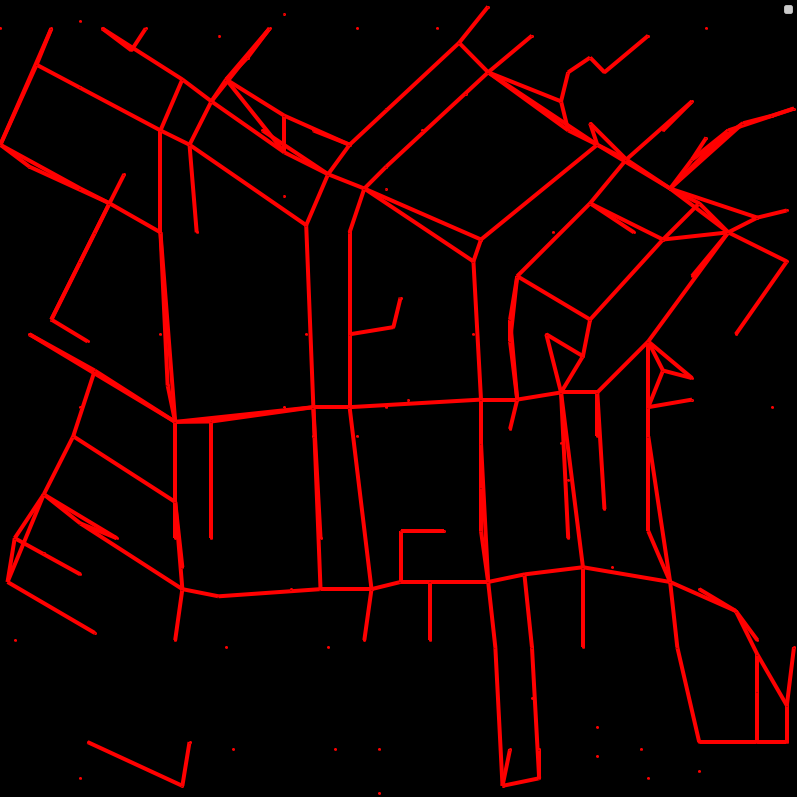} &
		\includegraphics[width=0.30\textwidth]{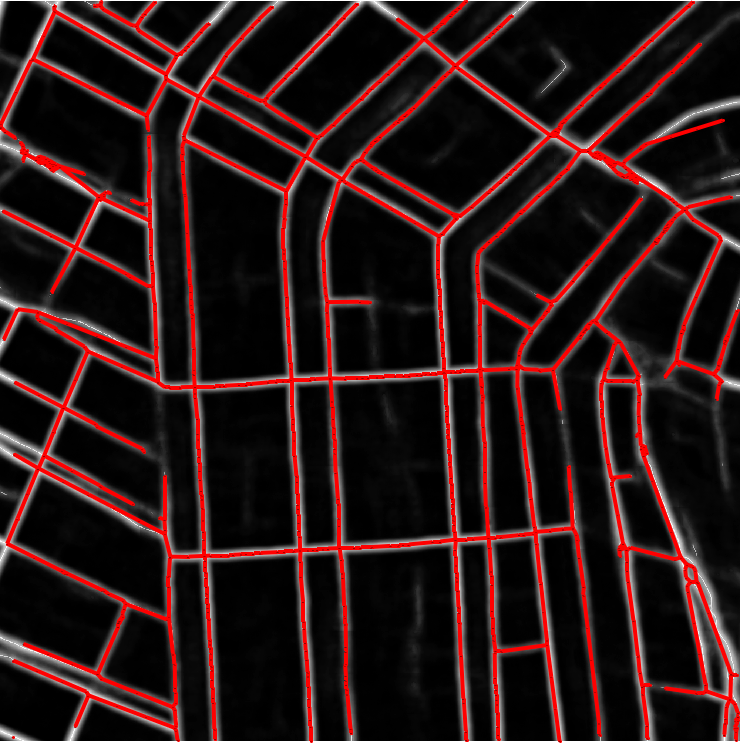} &
		\includegraphics[width=0.30\textwidth]{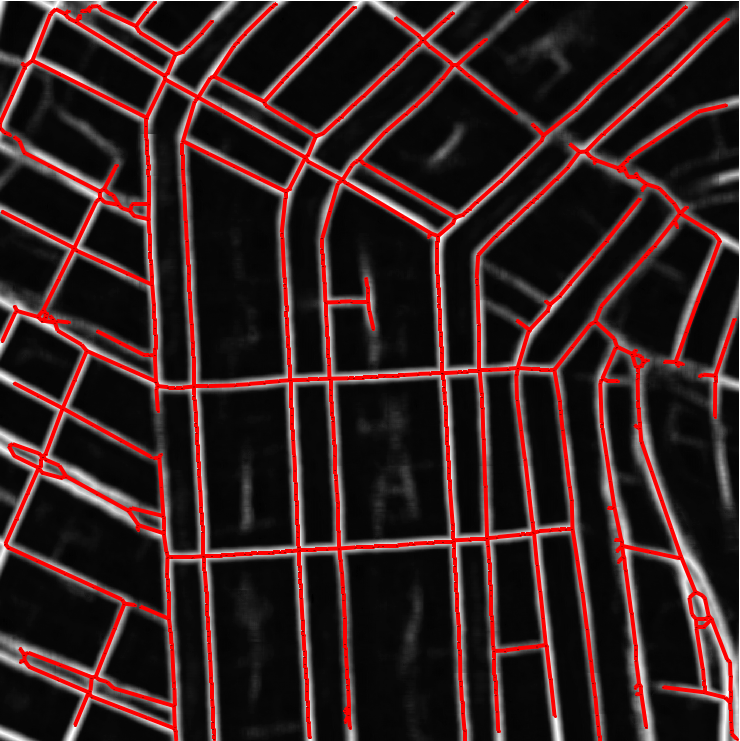} \\
		\PolyM{} & 
		\UNet{}+\OursGlobal{}& 
		\UNet{}+\OursWindowed{}\\		
	\end{tabular}
	\caption{
		Comparative results on the \RTracer{} dataset.
                For our results, we overlaid the graphs on the inferred distance maps.
		\label{fig:results-roadtracer}
	}
\end{figure*}

\begin{figure*}
\centering
\setlength{\tabcolsep}{1pt}
\begin{tabular}{@{} c c c c @{}}
\multicolumn{4}{c}{
  \includegraphics[width=0.32\textwidth]{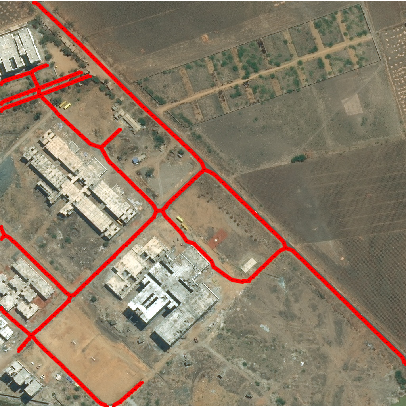} 
  \includegraphics[width=0.32\textwidth]{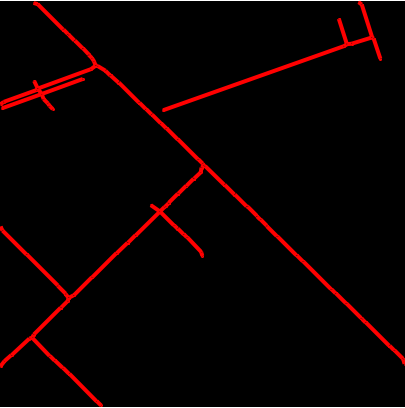} 
  \includegraphics[width=0.32\textwidth]{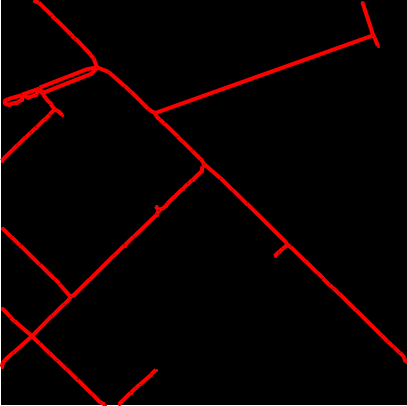} 
}
\\
\multicolumn{4}{c}{
  \makebox[0.32\textwidth][c]{\it input}  
  \makebox[0.32\textwidth][c]{\LinkN{}} 
  \makebox[0.32\textwidth][c]{\MultiB{}}  
}
 \\ 
\includegraphics[width=0.24\textwidth]{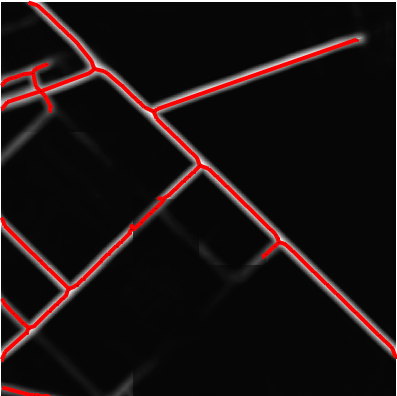} &
\includegraphics[width=0.24\textwidth]{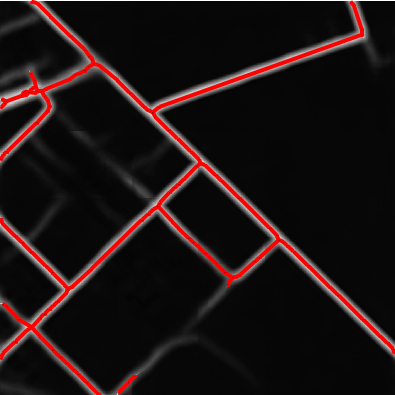} &
\includegraphics[width=0.24\textwidth]{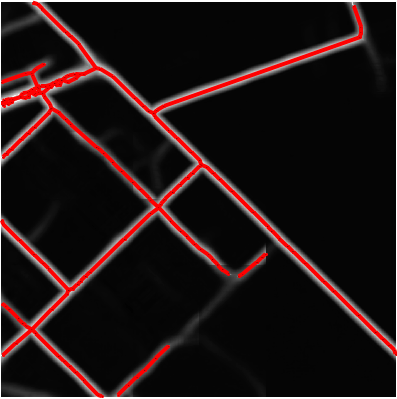} &
\includegraphics[width=0.24\textwidth]{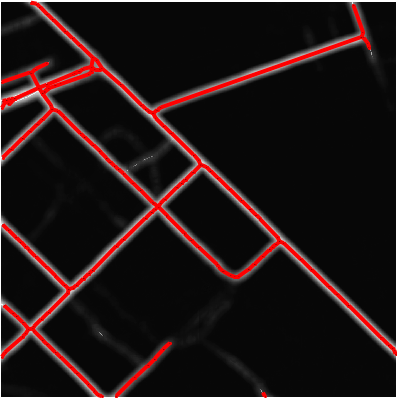} \\
\UNet{}  & 
\Dru{} & 
\UNet{}+\OursWindowed{} &
\Dru{}+\OursWindowed{} \\ 
\end{tabular}

\caption{
Comparative results on the \DG{} dataset.
For the results of our method, we overlaid graphs on the inferred distance maps.
\label{fig:results-deepglobe}
}
\end{figure*}

\begin{figure*}[!htb]
\centering
\setlength{\tabcolsep}{1pt}
\begin{tabular}{@{} c c c @{}}
	\multicolumn{3}{c}{
		\includegraphics[width=0.32\textwidth]{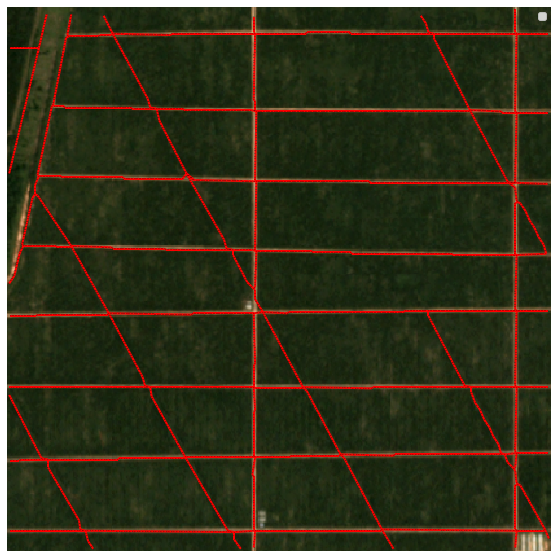} 
		\includegraphics[width=0.32\textwidth]{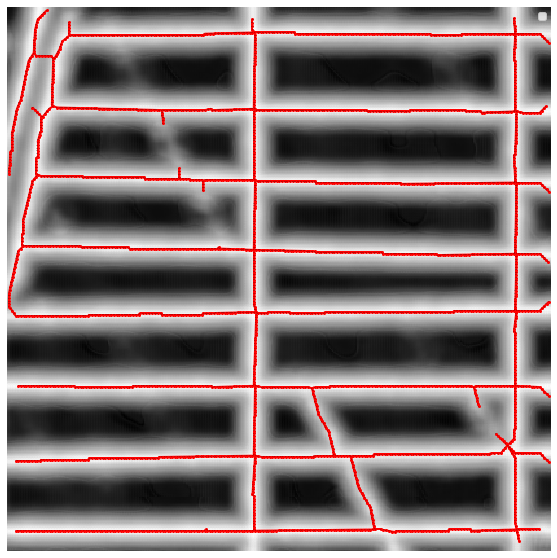} 
		\includegraphics[width=0.32\textwidth]{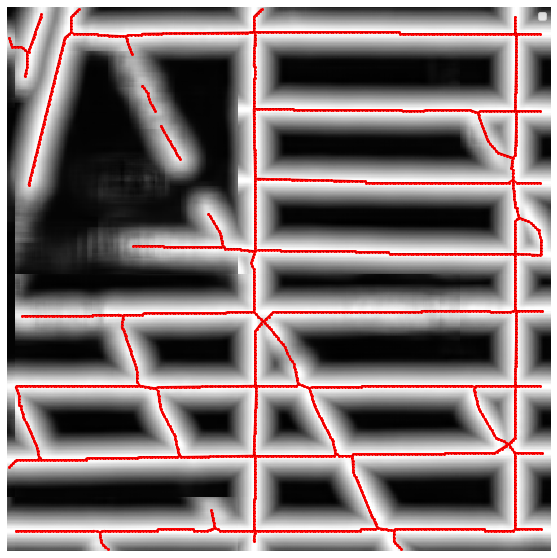} 
	}
	\\
	\multicolumn{3}{c}{
		\makebox[0.32\textwidth][c]{\it input}  
		\makebox[0.32\textwidth][c]{\UNet{}} 
		\makebox[0.32\textwidth][c]{\Dru{}}  
	}
	\\ 
	\includegraphics[width=0.484\textwidth]{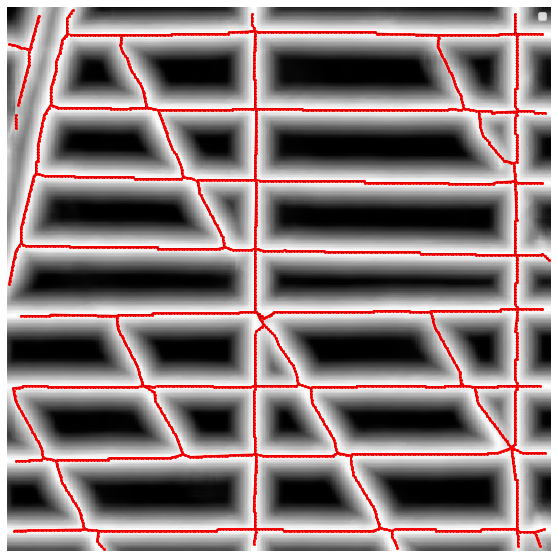} &
	\includegraphics[width=0.484\textwidth]{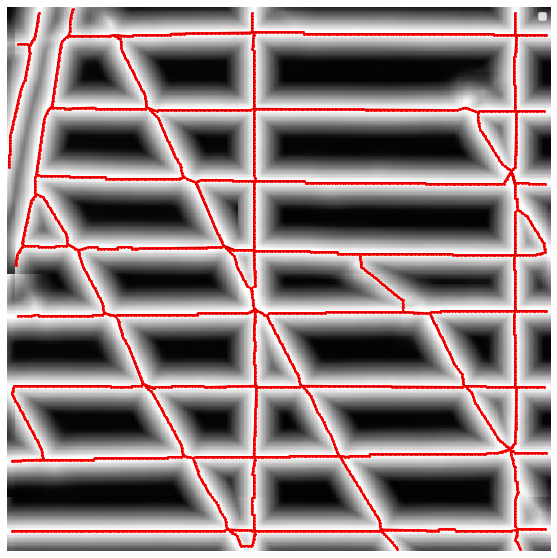} \\
	\UNet{}+\OursWindowed{} &
	\Dru{}+\OursWindowed{} \\ 
\end{tabular}

\caption{
Comparative results on the \CNL{} dataset.
For the results of our method, we overlaid graphs on the inferred distance maps.
\label{fig:results-canals}
}
\end{figure*}

We report the performance of our method on the \RTracer{}, \DG{} and \CNL{} datasets in Tabs.~\ref{tab:results-roadtracer},~\ref{tab:results-deepglobe} and~\ref{tab:results-canals}.
These corresponding network reconstructions are depicted qualitatively in Figs.~\ref{fig:results-roadtracer},~\ref{fig:results-deepglobe} and~\ref{fig:results-canals}. 
On average, \DG{} features simpler roads with fewer opportunities for mistakes than \RTracer{} and \CNL{}. Yet, in all datasets, we can capture connectivity more reliably than competing methods. 

Using the windowed version of our loss function boosts the performance of a simple U-Net past that of {\it all} the baselines on {\it all measures}, except CCQ on the \DG{} dataset.
The last is not surprising because our loss is designed to enforce connectivity, which CCQ does not measure. 
What is remarkable is that we were able to achieve this result using the comparatively simple U-Net architecture, whereas many of the competing architectures are far more sophisticated.
When the network is more powerful, our loss function boosts its performance even further.
As can be seen in Tab.~\ref{tab:results-deepglobe}, \Dru{} yields results as good as \MultiB{}, the best performer on the \DG{} dataset, already when trained with the mean squared error. 
Training \Dru{} with our loss increases its performance in terms of {\it all the scores}. 
Thanks to its increased stability, the windowed version of the loss slightly outperforms the global one.
The one exception is the \Junc{} measure computed on the \RTracer{} dataset, where the global version performs better than the windowed one.
We attribute this to the slightly increased tendency of the network to create road bifurcations when using the windowed-loss, which has little effect on the other metrics. 

\subsection{Ablation Studies}
We run a number of ablation studies to investigate the impact of the hyper-parameters of our method on performance.

\subsubsection{Varying the impact of the connectivity loss} 
\begin{figure*}[!htb]
\centering
\begin{tabular}{@{} >{\centering\arraybackslash}m{0.31\textwidth} >{\centering\arraybackslash}m{0.31\textwidth} >{\centering\arraybackslash}m{0.31\textwidth} @{}}
\includegraphics[width=0.31\textwidth]{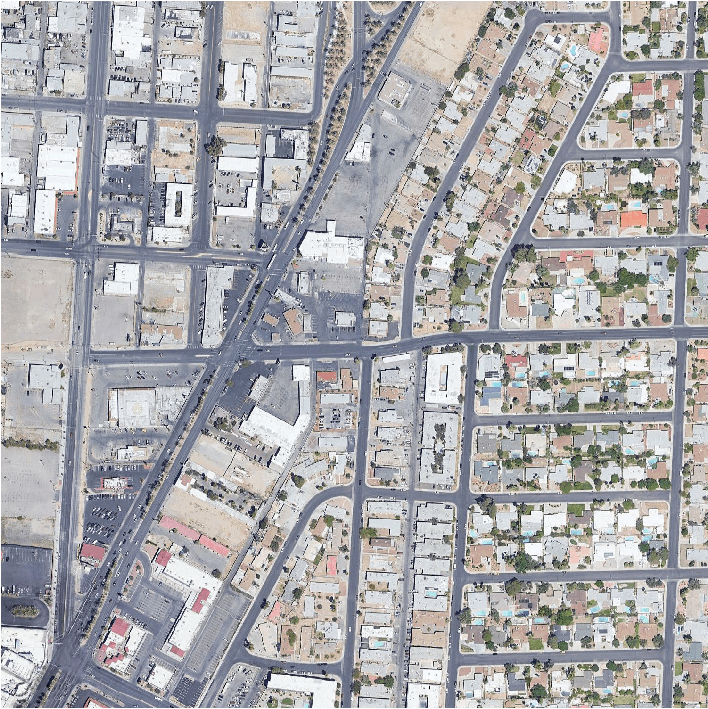} &
\includegraphics[width=0.31\textwidth]{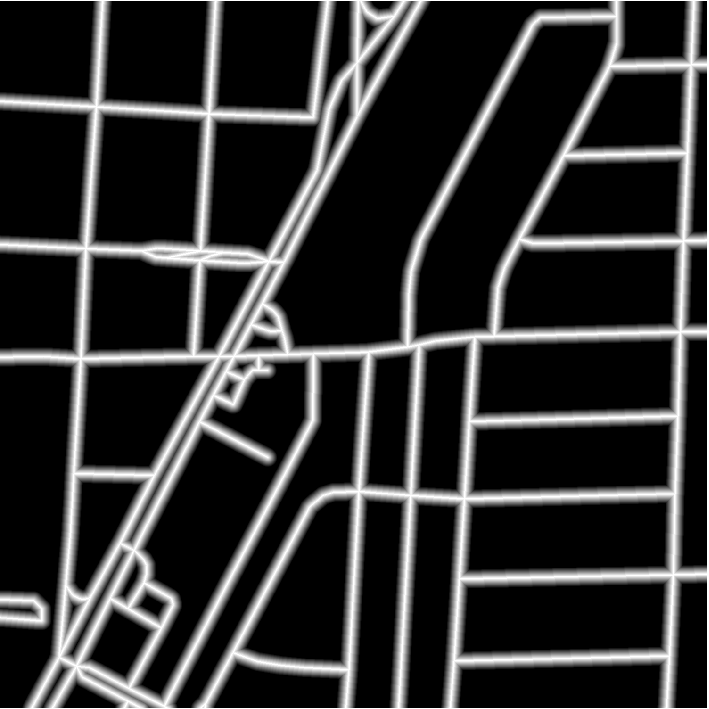} &
\includegraphics[width=0.31\textwidth]{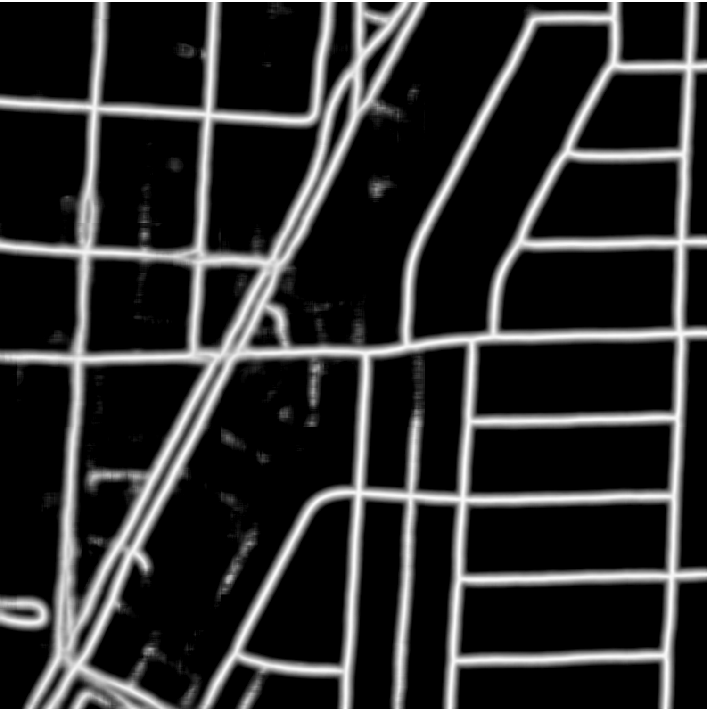} \\
{\it input}  & 
{\it ground-truth}& 
$\alpha=1e-5$ \\
	\includegraphics[width=0.31\textwidth]{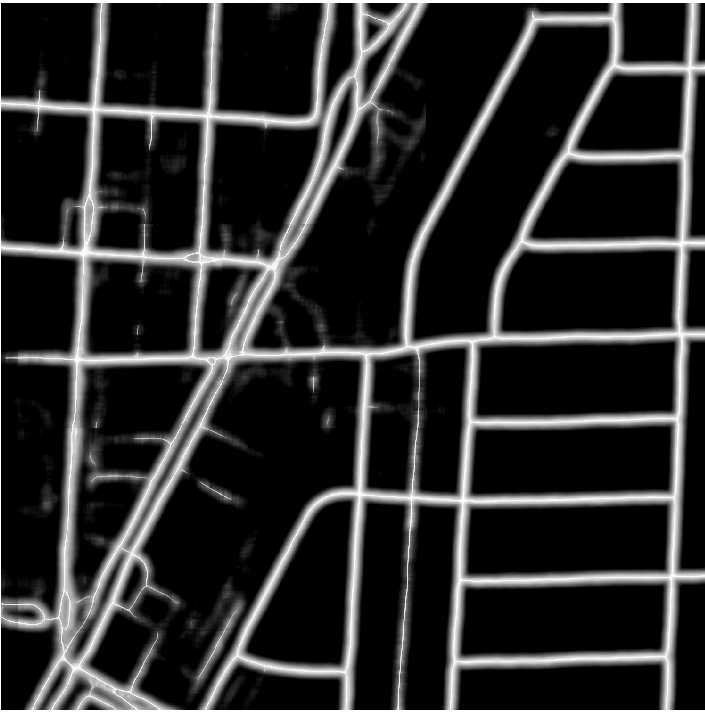} &
	\includegraphics[width=0.31\textwidth]{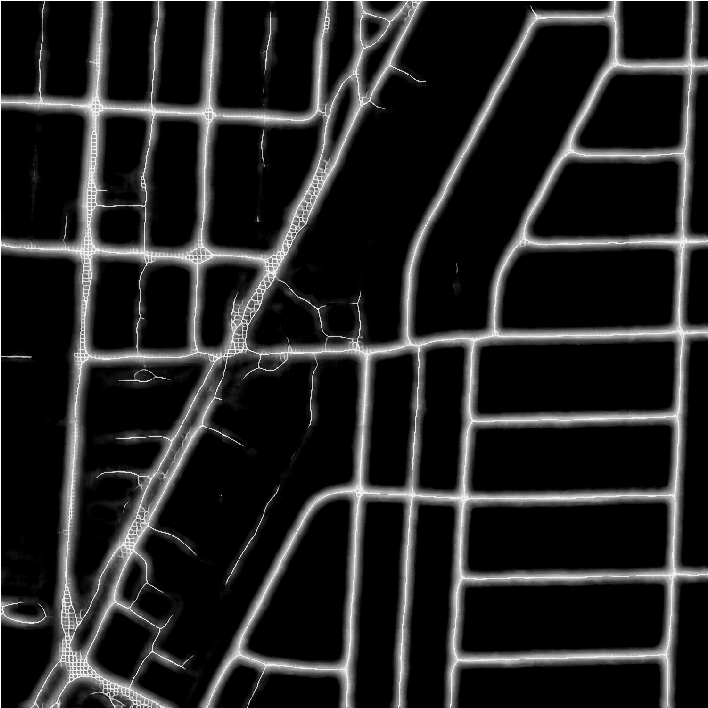} &
	\includegraphics[width=0.31\textwidth]{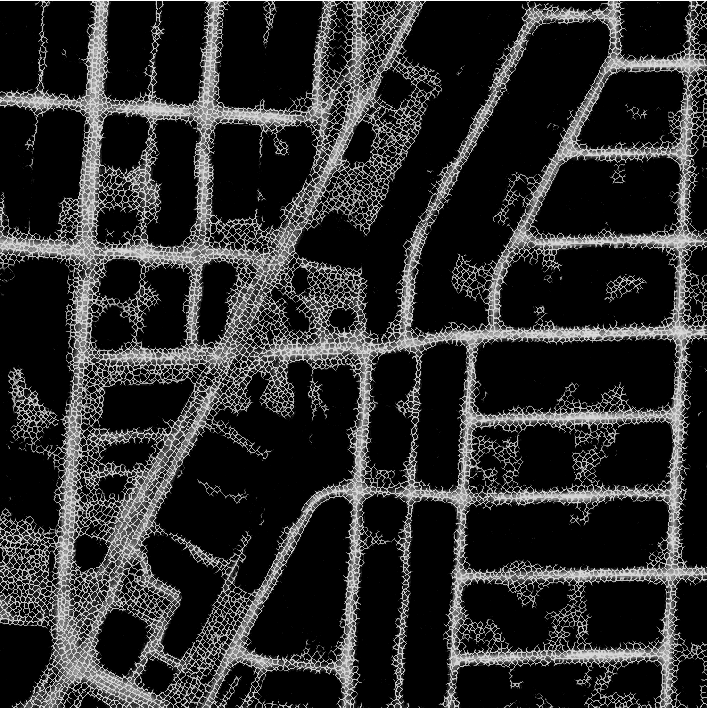} \\
	$\alpha=1e-4$ &
	$\alpha=1e-3$ & 
	$\alpha=1e-2$ \\
\end{tabular}

\vspace{-3mm}
\caption{
Effect of $\alpha$ on the distance map output by the neural network. 
As $\alpha$ is increased, the road map becomes more complete.
However, high values of $\alpha$ promote creating erroneous connections even where no roads are present in the image.
The corresponding numerical results can be found in Tab.~\ref{tab:results-malislr}.
\label{fig:visual-malislr}
}
\end{figure*}

As defined in Eq.~(2), our loss function is a combination of the mean square error with the connectivity term $\LTOPO$. 
The influence of the connectivity term on the loss is controlled by coefficient $\alpha$. 
We varied $\alpha$ to investigate its impact on the distance maps produced by the network.
The results presented in Tab.~\ref{tab:results-malislr} show that setting this coefficient too-low or too-high adversely affected performance, and its optimal value is in the order of $1e-4$.
The explanation of this phenomenon is provided in Fig.~\ref{fig:visual-malislr}.
For low values of $\alpha$, the effect of the connectivity-oriented component of the loss function is negligible.
When $\alpha$ is increased, more and more connections are represented in the distance map.
However, when $\alpha$ is set very high, the network starts to privilege disconnecting background image regions, even with no obvious roads in the input, creating false positive road segments.
\begin{table}[t]
	\centering
	\caption{
The impact of changing the $\alpha$ coefficient, balancing $\LMSE$ and $\LTOPO$, on performance.
Results of experiments on the \RTracer{} dataset.
Window size is fixed to 64x64 and $\beta$ to 0.1.
\UNet+\OursWindowed{} is used in all experiments.
Visualization of the corresponding results can be found in Fig.~\ref{fig:visual-malislr}.
\label{tab:results-malislr}
	}
	\begin{tabular}{@{} p {0.29\columnwidth} >{\centering\arraybackslash}p{0.7cm}>{\centering\arraybackslash}p{0.5cm}>{\centering\arraybackslash}p{0.5cm} >{\centering\arraybackslash}p{0.4cm} >{\centering\arraybackslash}p{0.002cm} >{\centering\arraybackslash}p{1.45cm} @{} }
		\cmidrule{2-7}
		
		& \multicolumn{4}{c}{Connectivity-oriented} && pixel-based \\
		
		\cmidrule{2-5}
		\cmidrule{7-7}
		
		$\alpha$ &    \APLS{} &       \TLTS  &      \Junc &         \HM &&        \CCQ \\
		\cmidrule{1-7}
		\text{ 1e-3} &
		72.3 &       46.3  &       80.3  &       73.1  &&  66.5 \\
		\text{ 1e-4} &
		\textbf{75.8} &      \textbf{49.7}  &       \textbf{82.8}  &       \textbf{76.0}  && \textbf{68.6}\\
		\text{ 1e-5} &   71.4    &       45.9     &  81.9  &       73.4      &&    67.1   \\
		\text{ 0.0} &
		66.3 &       40.0  &       77.5  &       68.2  &&  59.3 \\
		\cmidrule{1-7}
	\end{tabular}
	
\end{table}

\subsubsection{Balancing connectivity versus dis-connectivity} 
We counteract the tendency to produce excessively connected road networks by incorporating into the connectivity term a component penalizing false positive road segments, in addition to the one encouraging connectivity of true roads.
As specified in Eq.~(7), the coefficient $\beta$ balances these two terms.
We varied $\beta$ to investigate its impact on performance.
We present the results in Tab.~\ref{tab:results-beta}.
According to all the performance measures, the best results are obtained for $\beta=0.1$, meaning that the term preventing disconnections should have ten times more impact on the loss than the term preventing false positive roads.
\begin{table}[t]
	\centering
	\caption{
The impact of $\beta$, balancing the connectivity and dis-connectivity components of our loss, on performance.
Results of experiments on the \RTracer{} dataset.
\UNet+\OursWindowed{} is used in all experiments.
\label{tab:results-beta}
	}
	\begin{tabular}{@{} p {0.29\columnwidth} >{\centering\arraybackslash}p{0.7cm}>{\centering\arraybackslash}p{0.5cm}>{\centering\arraybackslash}p{0.5cm} >{\centering\arraybackslash}p{0.4cm} >{\centering\arraybackslash}p{0.002cm} >{\centering\arraybackslash}p{1.45cm} @{} }
		\cmidrule{2-7}
		
		& \multicolumn{4}{c}{Connectivity-oriented} && pixel-based \\
		
		\cmidrule{2-5}
		\cmidrule{7-7}
		
		$\beta$ &    \APLS{} &       \TLTS  &      \Junc &         \HM &&        \CCQ \\
		\cmidrule{1-7}
		\text{ 1e-0} &
		 71.4    &    43.8     &    80.9     &  73.2  && 66.5\\
		\text{ 1e-1} &
		\textbf{75.8} &      \textbf{49.7}  &       \textbf{82.8}  &       \textbf{76.0}  && \textbf{68.6}\\
		\text{ 1e-2} &    74.3     &      46.2      &   79.5   &    74.5     &&    65.3   \\
		\cmidrule{1-7}
	\end{tabular}
	
\end{table}

\subsubsection{Varying the window size} 
The third, and last, hyper-parameter of our method is the window size.
Computing the loss in windows, or image crops, as opposed to globally in the entire image, has the advantage of preventing accumulating all the error signal in a single pixel. 
The smaller the window, the more evenly the gradient is distributed among road pixels. 
The windowed version of the loss also enables enforcing connectivity of dead-ending roads, as small windows are often subdivided even by dead-ending roads.
Large window sizes do not have this effect, as roads shorter than the window size end in the middle of the window, without splitting it into disjoint tiles.
To discover the optimal window size, we tested its effect on performance. 
The results, presented in Tab~\ref{tab:results-window-rt} and~\ref{tab:results-window}, confirm that mid-size windows work best.
Setting the window size to $64\times64$ pixels resulted in the highest performance, and increasing or decreasing the window decreases performance.
\begin{table}[!htb]
\centering
\caption{
The impact of window size on performance.
Results of experiments on the \RTracer{} dataset.
$\alpha$ is fixed to $1e-4$ and $\beta$ to $0.1$.
\UNet+\OursWindowed{} is used in all experiments.
\label{tab:results-window-rt}
}
\begin{tabular}{@{} p {0.29\columnwidth} >{\centering\arraybackslash}p{0.7cm}>{\centering\arraybackslash}p{0.5cm}>{\centering\arraybackslash}p{0.5cm} >{\centering\arraybackslash}p{0.4cm} >{\centering\arraybackslash}p{0.002cm} >{\centering\arraybackslash}p{1.45cm} @{} }
\cmidrule{2-7}

& \multicolumn{4}{c}{Connectivity-oriented} && pixel-based \\

\cmidrule{2-5}
\cmidrule{7-7}

 {\it Window Size} &    \APLS{} &       \TLTS  &      \Junc &         \HM &&        \CCQ \\
\cmidrule{1-7}
\text{ (16x16)} &
			68.3 &      39.2  &       79.2  &       67.4  && 59.4\\
\text{ (32x32)} &
            72.1 &      45.8  &       78.9  &       72.7  && 65.7\\
\text{ (64x64)} &
        	75.8 &      \textbf{49.7}  &       \textbf{82.8}  &       \textbf{76.0}  && \textbf{68.6}\\
\text{ (128x128)} &     
			\textbf{76.1} &      46.4  &       81.7  &       74.5  && 68.3\\
\cmidrule{1-7}
\end{tabular}

\end{table}

\begin{table}[!htb]
\centering
\caption{
The impact of changing the window size on performance.
Results of experiments on the \DG{} dataset.
$\alpha$ is fixed to $1e-4$ and $\beta$ to $0.1$.
\UNet+\OursWindowed{} is used in all experiments.
\label{tab:results-window}
}
\begin{tabular}{@{} p {0.29\columnwidth} >{\centering\arraybackslash}p{0.7cm}>{\centering\arraybackslash}p{0.5cm}>{\centering\arraybackslash}p{0.5cm} >{\centering\arraybackslash}p{0.4cm} >{\centering\arraybackslash}p{0.002cm} >{\centering\arraybackslash}p{1.45cm} @{} }
	\cmidrule{2-7}
	
& \multicolumn{4}{c}{Connectivity-oriented} && pixel-based \\

\cmidrule{2-5}
\cmidrule{7-7}

{\it Window Size} &    \APLS{} &       \TLTS  &      \Junc &         \HM &&        \CCQ \\
\cmidrule{1-7}
\text{ (32x32)} &
            74.1 &       67.3  &       65.2  &       73.4  &&         74.8 \\
\text{ (64x64)} &
        \textbf{75.2} &      \textbf{69.8}  &       71.2  &       \textbf{79.8}  && 77.0\\
\text{ (128x128)} &     74.3    &       68.2     &  \textbf{72.0}  &       79.6      &&     \textbf{77.2}   \\
\cmidrule{1-7}
\end{tabular}

\end{table}

\subsubsection{Comparing Mean Squared Error to Cross Entropy}
Our loss function combines a connectivity-oriented term with mean squared error.
This combination outperforms a number of existing networks, trained with cross entropy.
We therefore investigated if just switching from the more common cross entropy to mean squared error, without our connectivity-oriented loss, impacts the performance. We present the results in Tab.~\ref{tab:results-roadtracer-ce}. 
We conclude that solely switching from pixel classification to distance map estimation does not warrant the increased connectivity, and its the addition of our connectivity-oriented term that does it.
\begin{table}[t]
	\centering
	\caption{
Comparison of Cross Entropy and Mean Square Error.
Results of experiments on the \RTracer{} dataset~\cite{Bastani18}.
\label{tab:results-roadtracer-ce}
	}
	\begin{tabular}{@{} p {0.335\columnwidth} >{\centering\arraybackslash}p{0.7cm}>{\centering\arraybackslash}p{0.5cm}>{\centering\arraybackslash}p{0.5cm} >{\centering\arraybackslash}p{0.4cm} >{\centering\arraybackslash}p{0.002cm} >{\centering\arraybackslash}p{1.45cm} @{} }
		\cmidrule{2-7}
		
		& \multicolumn{4}{c}{Connectivity-oriented} && pixel-based \\
		
		\cmidrule{2-5}
		\cmidrule{7-7}
		
		Method &    \APLS{} &       \TLTS &      \Junc &         \HM &&        \CCQ \\
		\cmidrule{1-7}
		\UNet{}{\it-CE}~\cite{Ronneberger15}&
		60.4 &        30.6 &       79.2 &        74.2 &&        63.3 \\ 
		\UNet{}{\it-MSE}~\cite{Ronneberger15}&
		66.3 &        40.0 &        77.5 &       68.2 &&        59.3 \\ 
		\cmidrule{1-7}
		\UNet+\OursGlobal{}   &      72.5 &        46.3 & \textbf{84.7}&       70.3 &&        63.8 \\
		\UNet+\OursWindowed{} & \textbf{75.8} & \textbf{49.7} &        82.8 &\textbf{76.0} && \textbf{68.6}\\
		\cmidrule{1-7}
	\end{tabular}
\end{table}

\section{Conclusion and future work}

We have introduced a differentiable loss function that effectively enforces proper connectivity on the output of binary segmentation ConvNets for the purpose of road network delineation.
Using this loss function to train a simple U-Net allows us to outperform far more sophisticated architectures on challenging benchmark datasets.
This suggests that we may not yet have unleashed the full power of these simpler networks and that adding appropriate constraints during training might be a way to do so. 

We have so far limited ourselves to networks of roads and drainage canals, but networks of linear structures are also pervasive in biomedical 3D imagery.
They range from neural structures to blood vessels and many others.
In future work, we will therefore expand our approach to handle 3D image stacks and address a much broader range of applications.

\bibliographystyle{IEEEtran}
\bibliography{string,biomed,vision,graphics,learning,misc}
\end{document}